\documentclass[11pt]{article}
\usepackage{acl}
\usepackage{booktabs}
\usepackage{array}
\usepackage{makecell}
\usepackage{times}
\usepackage{latexsym}
\usepackage{amsmath} 
\usepackage{algorithm}
\usepackage{algpseudocode}
\usepackage{placeins}
\usepackage[table]{xcolor}
\usepackage[most]{tcolorbox}
\usepackage{tabularx}
\usepackage{xcolor}
\usepackage{graphicx}
\usepackage{siunitx}
\usepackage{threeparttable}
\usepackage{listings}
\usepackage{xcolor}
\usepackage{xcolor}

\lstdefinelanguage{json}{
    basicstyle=\ttfamily\normalsize,
    string=[s]{"}{"},
    stringstyle=\color{black},
    comment=[l]{//},
    commentstyle=\color{gray},
    showstringspaces=false,
    breaklines=true,
    breakatwhitespace=true,
    columns=fullflexible,
    keepspaces=true,
    tabsize=2
}

\usepackage[T1]{fontenc}
\usepackage{adjustbox}
\usepackage[utf8]{inputenc}
\usepackage{microtype}
\usepackage{inconsolata}
\usepackage[most]{tcolorbox}
\tcbuselibrary{breakable}

\usepackage{multirow}

\definecolor{bestgreen}{HTML}{6FBE73}
\definecolor{secondgreen}{HTML}{D8F2D2}

\newcommand{\best}[1]{\cellcolor{bestgreen}#1}
\newcommand{\secondg}[1]{\cellcolor{secondgreen}#1}

\title{Schema-Guided LLM Verification and Conflict Resolution between LLM-Based and Data-Driven Causal Discovery }

\author{
\textbf{Amana Batool\textsuperscript{1}},
\textbf{Zhangchi Qiu\textsuperscript{1}},
\textbf{Christopher P. Carty\textsuperscript{2}} \\
\textbf{Michael Schwartz\textsuperscript{3}},
\textbf{Alan Wee-Chung Liew\textsuperscript{1}}\thanks{Corresponding author.},\\
\textsuperscript{1}School of Information and Communication Technology\\
\textsuperscript{2}School of Medicine and Dentistry \\
Griffith University, Australia \\
\textsuperscript{3}Department of Orthopedic Surgery,
University of Minnesota, USA \\
\texttt{amana.batool@griffithuni.edu.au}
\quad \texttt{a.liew@griffith.edu.au}
}

\begin{document}
\maketitle

\begin{abstract}
Learning causal structure from observational data requires more than statistical association. Data-driven causal discovery methods can recover unsupported edges when their assumptions are violated, or samples are limited. Large language models (LLMs) can interpret variable semantics but may produce judgments unsupported by the data. We introduce LLM-based conflict resolution (LLM-CR), a hybrid causal discovery framework that reconciles these complementary sources of evidence. LLM-CR separately constructs a schema-guided, verified LLM graph and a data-driven causal graph, retains agreed edges, and evaluates disputed directions using an uncertainty-aware conflict-resolution score that combines statistical fit, nonlinear dependence evidence, and an LLM prior calibrated by semantic entropy.
It further revisits plausible missing edges while preserving acyclicity. We evaluate LLM-CR on five primary causal-structure benchmarks, one auxiliary benchmark treated as potentially contaminated (Sachs), and one clinical prediction dataset derived from MIMIC-IV. Extensive experiments show that LLM-CR generally improves causal structure recovery over data-driven, LLM-only, and hybrid baselines, while producing useful causal graphs for downstream clinical prediction. The source code and publicly shareable datasets are available at \url{https://github.com/Amana-Batool/Framework-LLM-CR}.
\end{abstract}
\section{Introduction} 
Causal discovery aims to recover directed causal structures from observational data~\cite{glymour2019review} when randomized controlled trials are limited or infeasible because of ethical constraints~\citep{nichol2010challenging,steeger2021common}. Traditional data-driven causal discovery methods are grounded in principles from information theory and graphical models~\citep{zanga2022survey}, but they are vulnerable to small or noisy samples,  Markov equivalence ambiguity~\cite{mooij2020constraint}, and distributional assumptions~\citep{zhou2022survey}.
Crucially, these methods discover edges based on statistical dependence alone and have no mechanism for evaluating whether a recovered edge is semantically meaningful in the domain~\cite{du2025causal,peng2026causcientist}, a limitation that motivates incorporating external semantic knowledge into the discovery process.

\begin{figure}[t]
  \centering
  \includegraphics[width=\columnwidth]{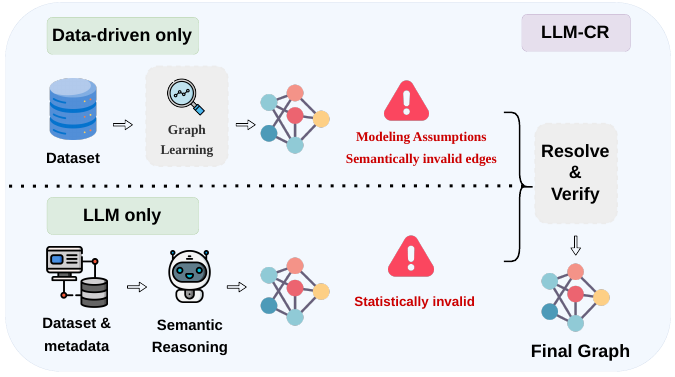}
  \caption{Conceptual comparison of causal discovery approaches. Data-driven methods emphasize statistical evidence, whereas LLM-only methods emphasize semantic plausibility. Our framework resolves their disagreements through schema-guided verification and uncertainty-aware conflict resolution.}
  ~\label{Figure1}
\end{figure}

Large language models (LLMs) have attracted significant interest as a knowledge source for causal discovery, owing to their ability to encode broad world knowledge and reason over variable semantics acquired during pretraining~\citep{kiciman2024causal,ban2025integrating,du2025causal}. 
Prior work has therefore used LLMs to directly predict pairwise causal relations~\cite{kiciman2024causal}, infer causal orderings~\cite{vashishtha2025causal}, and answer commonsense causal
questions~\cite{chi2024unveiling}.
Despite these advances, LLM-only approaches still face key limitations. First, LLMs behave as unconstrained reasoners that do not enforce structural validity and are prone to overconfidence and hallucination~\cite{xiong2024can,zhang2025siren,du2025causal}. Second, they generate graphs that lack statistical validation~\citep{kiciman2024causal,zevcevic2023causal}, making them unreliable as standalone causal decision-makers.

These limitations motivate combining LLM semantic knowledge with data-driven statistical evidence. Some approaches use LLMs as soft priors that reduce the hypothesis space for data-driven search~\cite{darvariu2024large,
ban2025integrating}, yet these priors are applied without constraining variable roles, semantic categories, or forbidden edge directions in the LLM output.
Other methods integrate LLM reasoning into a data-driven discovery pipeline, either as edge-orientation assistants for resolving Markov equivalence ambiguity~\cite{long2023causal,vashishtha2025causal}, or as iterative augmenters that combine LLM proposals with traditional discovery algorithms such as PC~\cite{khatibi2024alcm,du2025causal}.
More recently, CauScientist~\cite{peng2026causcientist} has grounded LLM-proposed graph
modifications in observed data by accepting modifications that improve a whole-graph BIC score. However, it does not constrain the LLM judgments at their source, combine complementary statistical evidence, or calibrate
per-edge LLM confidence when balancing semantic and statistical evidence.

To address these limitations, we propose LLM-based conflict resolution (\textsc{LLM-CR}), a hybrid causal discovery framework that treats data-driven and LLM-derived graphs as complementary evidence sources.
The framework first uses schema-guided LLM reasoning to generate and verify pairwise causal judgments under semantic and directional constraints. In parallel, a data-driven method estimates a graph from observational data. \textsc{LLM-CR} then unifies the two graphs and applies an uncertainty-aware conflict-resolution score composed of three components: BIC score, HSIC score, and an uncertainty-aware LLM prior. Score-based causal discovery methods such as GES commonly use the Bayesian Information Criterion (BIC) to balance model fit against graph complexity \citep{chickering2002optimal,nazaret2024extremely}. Kernel-based methods employ the Hilbert--Schmidt Independence Criterion (HSIC) to detect nonlinear and non-Gaussian dependencies and to assess residual independence in causal direction discovery \citep{gretton2007kernel,hoyer2008nonlinear,peters2014causal}. However, BIC alone may be sensitive to model assumptions, while HSIC does not explicitly penalize structural complexity or incorporate semantic knowledge. LLM-CR therefore combines local BIC gain, HSIC-based dependence evidence, and an uncertainty-weighted LLM prior to resolve disputed edges using complementary statistical and semantic signals. Finally, a conservative edge-recovery step revisits plausible missing edges and retains only those that improve the hybrid score while preserving DAG validity. Figure~\ref{Figure1} conceptually contrasts data-driven, LLM-only, and LLM-CR causal discovery. 

We conduct extensive experiments across benchmark and clinical causal discovery settings, and the results show that \textsc{LLM-CR} more effectively balances statistical evidence with semantic causal knowledge, improving structural accuracy compared to data-driven and LLM-driven baselines on the benchmark datasets.
To summarize, our contributions are:
 \begin{itemize}
     \item We propose \textsc{LLM-CR}, a hybrid causal discovery framework that integrates schema-guided LLM reasoning with data-driven causal discovery to resolve disagreements between semantic and statistical evidence.
     \item We introduce a schema-guided LLM generation and verification module that guides LLM reasoning using semantic categories, causal roles, and forbidden-direction constraints, thereby reducing inconsistency and structurally invalid judgments.
     \item We develop an uncertainty-aware graph refinement mechanism that uses a hybrid score to resolve disputed edges and recover plausible missing edges while preserving DAG validity.
     \item Extensive experiments across five datasets demonstrate that \textsc{LLM-CR} consistently outperforms baselines on majority benchmarks.
 \end{itemize}

\section{Related Work}
\subsection{Data-driven Causal Discovery}
Current data-driven causal discovery methods include constraint-based, score-based, functional causal models, and optimization-based algorithms. Constraint-based methods like PC~\cite{spirtes2001causation} and FCI \citep{spirtes2013causal,colombo2012learning} use conditional independence testing and score-based methods like GES \cite{chickering2002optimal} and FGES~\cite{ramsey2017million} assign a score, such as likelihood or BIC, to each candidate graph and search for the graph with the best score \cite{squires2023causal}.
Either methods cannot distinguish graphs within the same Markov Equivalence classes \cite{zanga2022survey}. Functional causal methods such as additive noise models (ANMs) assume that noise is independent of inputs~\cite{peters2014causal}, which makes full graph recovery possible rather than Markov equivalent classes, but may fail in linear-Gaussian settings ~\cite{zhou2022survey}. Continuous optimization-based methods, such as DAGs with NOTEARS~\cite{zheng2018dags}, have transformed the combinatorial problem of DAG search into a continuous optimization problem, while a smooth function enforces acyclicity~\cite{zhou2022survey}. These methods reduce time complexity for causal discovery ~\cite{wang2024survey}, but their model flexibility remains the main limitation for complex non-linear data, and they struggle to enforce acyclicity efficiently~\cite{zhou2022survey}. Overall, data-driven methods provide a principled statistical framework for graph learning, but their performance is limited by sample size, noise, modeling assumptions, Markov equivalence classes, and a lack of semantic understanding of variables, making it hard to apply these methods to real-world systems~\cite{du2025causal}.
\subsection{LLM-based Causal Discovery} 
Recent studies have explored the use of LLMs for causal reasoning~\cite{jin2023cladder}, either as a standalone causal reasoner in causal discovery \cite{long2022can} or in combination with traditional data-driven causal discovery algorithms~\citep{ban2025llm, wang2025large} due to the promising performance of LLMs in causal discovery ~\citep{kiciman2024causal, abdulaal2024causal}. Some methods include prompting LLMs for a causal relationship between two or more variables, along with description and contextual knowledge of the variables ~\citep{long2023causal, darvariu2024large} and show that LLM can provide useful semantic and domain knowledge but is sensitive to prompts, may behave as imperfect experts, confuse semantic relatedness with direct causality and can infer many incorrect edges ~\citep{du2025causal, long2023causal, wan2025large} as~\cite{zevcevic2023causal} has stated LLMs as ``causal parrots" because LLMs may rely on textual correlation. Therefore, we propose a schema-guided approach for generating and verifying LLM responses, aiming to reduce the number of incorrect edges. Some studies have attempted to use LLMs to test conditional independencies and then apply them in constraint-based algorithms~\cite{cohrs2023large, du2025causal}, indicating growing interest in combining LLMs with traditional causal discovery algorithms. However, these methods rely heavily on LLMs to infer conditional independencies, which can vary widely. Some works have integrated LLMs as soft constraints, such as using Bayesian priors~\cite{ban2025integrating} or employing them as priors for causal discovery ~\cite{darvariu2024large}, or some studies have attempted to use LLMs for finding causal order of the variables instead of pairwise causal relationships~\cite{vashishtha2025causal}. Recently, statistically verified studies have improved reliability by incorporating statistical evidence into LLM-based suggestions, but statistical verification governs final acceptance~\cite{peng2026causcientist}. The above studies either use an LLM as an imperfect expert and apply consistency constraints such as acyclicity~\cite{long2023causal} or rely on statistical evidence for the final decision~\cite{peng2026causcientist}. This motivates our framework, which separates LLM reasoning and data-driven discovery into two branches and resolves their disagreement using a balanced approach based on an uncertainty-aware conflict-resolution score.
\begin{figure*}[t]
  \centering
  \includegraphics[width=\textwidth]{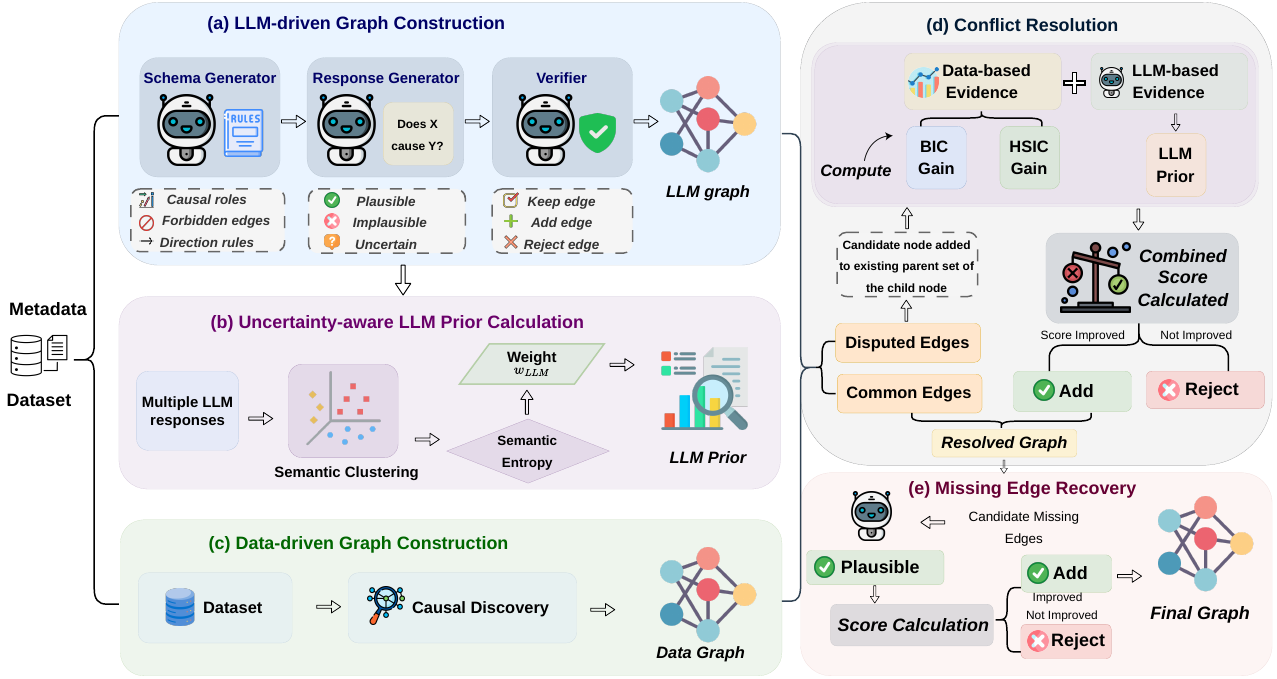}
  \caption{Overview of the proposed LLM-CR framework: (a) schema-guided LLM graph construction through schema generation, repeated causal judgments, and verification; (b) uncertainty-aware LLM prior estimation using semantic clustering, semantic entropy, and reliability weighting; (c) data-driven causal graph construction; (d) conflict resolution, which combines BIC gain, HSIC gain, and the uncertainty-aware LLM prior to evaluate disputed edges while retaining common edges; and (e) missing-edge recovery, which evaluates plausible omitted edges using the combined score and adds only score-improving edges to produce the final graph.}
  \label{fig:framework}
\end{figure*}
\section{Method}
\textsc{LLM-CR} treats LLM-derived and data-driven causal evidence as two independently constructed graphs over the same variables and reconciles their edge-level disagreement under an uncertainty-calibrated score. Rather than asking the score to repair an unconstrained LLM graph, we first make the LLM-derived graph auditable through schema-guided generation and verification, and only then resolve the remaining disagreements with the data-driven graph. Figure~\ref{fig:framework} shows the overall pipeline.

\begin{table}[t]
\centering
\small
\begin{tabular}{@{}p{0.22\columnwidth}p{0.72\columnwidth}@{}}
\toprule
\textbf{Stage} & \textbf{Representative prompts} \\
\midrule
Schema Generator &
Given the domain and variable metadata, generate a causal schema
and causal categories using dataset variables. Return strict JSON. \\
\midrule
Response Generator & Given metadata and the causal schema, assess whether the proposed direction $X\!\to\!Y$ is scientifically plausible; do not equate association with causation. \\
\midrule
Verifier & Given the schema, metadata, and prior majority decision for $X\!\to\!Y$, keep or correct that decision using a pair-specific causal rationale. \\
\bottomrule
\end{tabular}
\caption{Representative prompts for each stage during LLM-verified graph generation.}
\label{tab:prompt-excerpts}
\end{table}

\subsection{Problem Setup and Notation}
Given an observational dataset \(\mathcal{D}\) over variables \(\mathcal{V}\), causal discovery aims to estimate a DAG \(G=(\mathcal{V},\mathcal{E})\) where \(\mathcal{E}\)  represents edges that capture plausible direct causal relations among the observed variables \(\mathcal{V}\). We formulate this task as an uncertainty-aware graph unification problem between two imperfect evidence sources.  The objective is to construct a final acyclic graph that preserves edges supported by both evidence and resolves disputed edges using a combined score $S$.

The LLM-CR framework comprises four major stages: 1) \emph{Schema-guided LLM inference}, which produces a verified LLM graph $G_{\mathrm{LLM}}$; 2) \emph{Data-driven causal discovery},  which produces a statistical graph $G_{\mathrm{Data}}$; 3) \emph{Uncertainty-aware conflict resolution}, which resolves disputed edges using a combined score $S$; and 4) \emph{Missing edge recovery}, which adds plausible missing edges only when $S$ improves and acyclicity is preserved.

\subsection{Schema-Guided LLM Inference}
Prompting LLMs with variable names to infer causal edges is brittle, producing graphs sensitive to phrasing and devoid of structural guarantees. \textsc{LLM-CR} addresses this by decomposing LLM-based causal reasoning into three stages that progressively refine pairwise causal evidence under domain constraints. The \textbf{schema generator} first induces dataset-specific causal constraints from variable metadata. The \textbf{response generator} then queries the LLM for structured pairwise responses conditioned on these constraints. Finally, the \textbf{verifier} checks each aggregated judgment for consistency with the schema before admitting it into the LLM-derived graph~$G_{\mathrm{LLM}}$. Representative prompts are shown in Table~\ref{tab:prompt-excerpts}, and the Appendix provides detailed prompts and examples for each stage.

\subsubsection{Schema Generator}

Variable names and dataset descriptions are typically weakly structured metadata that offer limited causal guidance on their own.
Given the dataset variable names and metadata $M$, the schema generator uses the LLM to induce a dataset-specific causal schema~$C$ comprising three types of constraints.
(1)~\emph{causal role assignments} that classify each variable as a likely root cause, intermediate, or outcome; (2)~\emph{semantic categories} with category-level direction rules (e.g., demographic variables may cause clinical measurements but not vice versa); and (3)~\emph{forbidden-edge constraints} that exclude structurally inadmissible relations. The resulting schema~$C$ conditions all subsequent LLM queries and serves as the reference against which generated judgments are verified.

\subsubsection{Response Generator}
The response generator queries the LLM for structured pairwise causal evidence.
For each ordered variable pair~$(X_i,X_j)$, it receives the metadata~$M$ and schema~$C$ as context and samples $R$ independent responses. Each response is a structured record containing a categorical label $\in \{\textit{plausible},\textit{implausible},\textit{uncertain}\}$, a causal explanation, and token-level log-probabilities. These outputs serve two downstream purposes. 
The labels are aggregated and verified to construct~$G_{\mathrm{LLM}}$, while the explanations and log-probabilities are retained for uncertainty quantification via semantic entropy~\cite{farquhar2024detecting} as shown in Figure~\ref{fig:framework}(b).
\subsubsection{Verifier}
Category-level compatibility alone does not justify a specific edge; for instance, permitting demographic variables to influence clinical measurements would admit every such pair.
The verifier therefore re-examines each judgment and retains an edge only when a pair-specific causal mechanism supports it. The $R$ labels for each pair~$(X_i,X_j)$ are first aggregated via majority voting with threshold~$\tau_m$ to obtain a preliminary label.
The verifier then evaluates this label against the role assignments, direction rules, and forbidden-edge constraints in~$C$ and~$M$, returning one of four actions. \textit{Keep-edge} and \textit{add-edge} accept the directed relation, while \textit{remove-edge} and \textit{no-edge} reject it. 
The final graph $G_{\mathrm{LLM}}$ contains all pairs assigned \textit{keep-edge} or \textit{add-edge}. 
 
\subsection{Data-driven Causal Discovery}
In parallel with the LLM branch, a standard causal discovery algorithm is applied to the observational dataset~$\mathcal{D}$ to produce a data-driven graph $G_{\mathrm{Data}}$.
The framework is agnostic to the choice of algorithm and treats its output as complementary statistical evidence.
The resulting graph provides empirical support for candidate causal relations but may contain spurious or missing edges due to limited sample size, noisy measurements, or weak domain constraints.
Therefore, we treat this graph as statistical evidence and use it in the next stage to compare with the verified LLM-derived graph.

\subsection{Uncertainty-aware Conflict Resolution}
Once $G_{\mathrm{LLM}}$ and $G_{\mathrm{Data}}$ are constructed, 
the framework retains all edges on which the two graphs agree in a common-edge set $\mathcal{E}_{\mathrm{common}}$ and collects the remaining edges into a disputed set $\mathcal{E}_{\mathrm{disputed}}$. A dispute arises when an edge appears in only one graph or when the two graphs assign opposite directions to the same variable pair.

For each disputed edge $e_{ij}=X_i\rightarrow X_j$, the framework temporarily adds $X_i$ to the parent set of $X_j$ in $\mathcal{E}_{\mathrm{common}}$ and evaluates the candidate using three complementary signals. 
The local BIC gain~\cite{chickering2002optimal} measures whether the edge improves the conditional fit of $X_j$ enough to justify the additional model complexity. 
The HSIC-based~\cite{gretton2007kernel} residual dependence reduction measures whether introducing $X_i$ explains nonlinear dependence that remains in the residual of $X_j$ after conditioning on its current parents.
The uncertainty-aware LLM prior measures whether the proposed direction is semantically plausible and consistently supported across repeated LLM judgments. See Appendix~\ref{sec: appendix c} for computational details of each component.

For each candidate disputed edge \(e_{ij} = X_i \rightarrow X_j\), \textsc{LLM-CR} evaluates whether to accept or reject the edge based on improvement in score $S_{\mathrm{combined}}(e)$ as follows:
\begin{equation}
S_{\mathrm{combined}}(e) =
\lambda_{\mathrm{BIC}}(e)
+
\beta_{\mathrm{HSIC}}(e)
+
\gamma_{\mathrm{LLM}}(e)
\end{equation}
where $\lambda$, $\beta$, and $\gamma$ denote the coefficients that control the contribution of each component in the score. The component scores are standardized across all disputed candidate directions before aggregation so that no single component dominates due to scale differences alone.
A disputed edge is accepted only when its combined score is positive and adding it does not introduce a directed cycle. When $G_{\mathrm{Data}}$ and $G_{\mathrm{LLM}}$ assign opposite directions to the same unordered pair, both directions are scored, and the higher-scoring eligible direction is kept. The final resolved graph combines $\mathcal{E}_{\mathrm{common}}$ with all accepted disputed edges (Algorithm~\ref{alg:conflict-resolution}).

\subsection{Missing Edge Recovery}
After conflict resolution, the framework revisits directed edges that are absent from the resolved graph but receive strong LLM support from the response generator, defined as $p_{\mathrm{LLM}}(e)\geq 0.7$. For each such candidate $e=X_i\rightarrow X_j$, the same three-signal score $S_{\mathrm{combined}}(e)$ is recomputed. 
Candidates whose score exceeds the threshold $\tau_r$ are provisionally added to the graph. Because multiple recovered edges may jointly form a cycle, we subsequently repair the graph by repeatedly removing the lowest-scoring newly recovered edge from each detected cycle until the final graph is acyclic.

\section{Experiments}

\begin{table}[t]
\centering
\begin{tabular}{@{}lrrr@{}}
\toprule
\textbf{Dataset} & \textbf{Samples} & \textbf{Nodes} & \textbf{Edges} \\
\midrule
Coma      & 5,000 &  5 &  5 \\
Agro      & 5,000 &  6 & 10 \\
Gait      & 2,000&  8 & 16 \\
Alzheimer & 1,000&  9 & 17 \\
CP        & 2,157 & 11 & 30 \\
Sachs     & 7,463& 11 & 17 \\
MIMIC-IV  & 4,630 & 18 & -- \\
\bottomrule
\end{tabular}
\caption{Dataset statistics.}
\label{tab:dataset_statistics}
\vspace{-5pt}
\end{table}
Our investigation addresses the following research questions: \textbf{RQ1:} To what extent do LLMs show evidence of prior exposure to benchmark causal datasets? \textbf{RQ2:} How does LLM-CR’s performance compare to existing methods? \textbf{RQ3:} How useful are the recovered graphs for downstream clinical prediction? \textbf{RQ4:} What is the impact of each component in LLM-CR on overall performance? \textbf{RQ5:} How do scoring and recovery strategies affect graph quality?
\subsection{Experimental Setup}
\begin{table*}[t]
\centering
\small
\setlength{\tabcolsep}{4pt}

\resizebox{0.9\textwidth}{!}{%
\begin{tabular}{lccccc}
\toprule
\textbf{Method} & \textbf{CP} & \textbf{Gait} & \textbf{Coma} & \textbf{Agro} & \textbf{Alzheimer} \\
\midrule
PC & 0.183 $\pm$ 0.064 & 0.224 $\pm$ 0.036 & 0.571 $\pm$ 0.000 & 0.400 $\pm$ 0.000 & 0.089 $\pm$ 0.122 \\
GES & 0.502 $\pm$ 0.126 & \best{0.786 $\pm$ 0.000} & 0.571 $\pm$ 0.000 & 0.588 $\pm$ 0.000 & 0.335 $\pm$ 0.081 \\
PC-ANM & 0.233 $\pm$ 0.064 & 0.262 $\pm$ 0.048 & 0.607 $\pm$ 0.080 & 0.500 $\pm$ 0.000 & 0.087 $\pm$ 0.120 \\
Direct-LiNGAM & 0.334 $\pm$ 0.087 & 0.458 $\pm$ 0.071 & 0.756 $\pm$ 0.040 & 0.700 $\pm$ 0.000 & 0.270 $\pm$ 0.031 \\
NOTEARS & 0.067 $\pm$ 0.061 & 0.080 $\pm$ 0.110 & 0.000 $\pm$ 0.000 & 0.143 $\pm$ 0.000 & 0.091 $\pm$ 0.000 \\
DAG-GNN & 0.000 $\pm$ 0.000 & 0.021 $\pm$ 0.047 & 0.000 $\pm$ 0.000 & 0.170 $\pm$ 0.042 & 0.109 $\pm$ 0.038 \\
DAGMA-L & 0.182 $\pm$ 0.029 & 0.120 $\pm$ 0.110 & 0.571 $\pm$ 0.000 & 0.125 $\pm$ 0.000 & 0.033 $\pm$ 0.075 \\
\midrule
GPT-4o & 0.371 $\pm$ 0.044 & 0.176 $\pm$ 0.043 & 0.600 $\pm$ 0.000 & 0.073 $\pm$ 0.100 & 0.514 $\pm$ 0.035 \\
\quad + metadata & 0.539 $\pm$ 0.051 & 0.257 $\pm$ 0.030 & 0.667 $\pm$ 0.000 & \secondg{0.718 $\pm$ 0.042} & 0.673 $\pm$ 0.015 \\
Qwen3-8B & \secondg{0.567 $\pm$ 0.000} & 0.435 $\pm$ 0.000 & 0.500 $\pm$ 0.000 & 0.667 $\pm$ 0.000 & 0.111 $\pm$ 0.000 \\
\quad + metadata & 0.563 $\pm$ 0.000 & 0.370 $\pm$ 0.000 & \secondg{0.800 $\pm$ 0.000} & 0.700 $\pm$ 0.000 & \secondg{0.733 $\pm$ 0.027} \\
Qwen3-14B & 0.437 $\pm$ 0.044 & 0.111 $\pm$ 0.000 & 0.476 $\pm$ 0.130 & 0.000 $\pm$ 0.000 & 0.000 $\pm$ 0.000 \\
\quad + metadata & 0.430 $\pm$ 0.060 & 0.304 $\pm$ 0.040 & \secondg{0.800 $\pm$ 0.000} & 0.284 $\pm$ 0.110 & 0.091 $\pm$ 0.000 \\
\midrule

LLM-greedy & 0.169 $\pm$ 0.021 & 0.240 $\pm$ 0.000 & 0.480 $\pm$ 0.110 & 0.361 $\pm$ 0.061 & 0.013 $\pm$ 0.030 \\

LLM-CD & 0.436 $\pm$ 0.022 & 0.400 $\pm$ 0.050 & 0.250 $\pm$ 0.000 & 0.444 $\pm$ 0.000 & 0.608 $\pm$ 0.011 \\
\midrule
LLM-CR & \best{0.805 $\pm$ 0.007} & \secondg{0.683 $\pm$ 0.009} & \best{0.978 $\pm$ 0.050} & \best{0.871 $\pm$ 0.038} & \best{0.881 $\pm$ 0.013} \\
\bottomrule
\end{tabular}
}
\caption{F1 scores (mean $\pm$ standard deviation over five runs) for causal graph recovery on five datasets. \best{Dark green} and \secondg{light green} indicate the best and second-best mean F1 in each dataset respectively.}
\label{tab:all-datasets-f1}
\end{table*}

\paragraph{Datasets.} 
We evaluate LLM-CR on five datasets with known ground-truth causal structures, including Coma~\cite{cooper1984nestor}, Agro~\cite{baudrit2022decision}, Gait~\cite{steele2022causal}, Alzheimer~\cite{nagesh2026evaluating}, Cerebral Palsy (CP)~\cite{gill2023causal}, and on MIMIC-IV~\cite{du2025causal} for downstream clinical prediction, where no ground-truth causal graph is available, as shown in Table~\ref{tab:dataset_statistics}. Sachs~\cite{sachs2005causal} is reported separately as an auxiliary benchmark because the contamination analysis indicates a higher risk of prior exposure. See~\ref{appen_datsets} for details of the datasets.
\paragraph{Data contamination test. }
To assess whether the LLM may have memorized the causal structures during pretraining, we adopt the contamination-testing procedure of~\cite{babakov2025scalability} and report the predicted number of nodes and matched variable names for each dataset.
For each dataset, we prompt the LLM with the dataset name and paper title and ask it to produce the number of nodes and their names. Although exact node-name recall does not fully rule out indirect exposure to causal relations, it provides a practical proxy for assessing memorization risk.

\begin{table}[t]
\centering
\small
\setlength{\tabcolsep}{5pt}
\resizebox{1\linewidth}{!}{%
\begin{tabular}{@{}lcccc@{}}
\toprule
\multirow{2}{*}{\textbf{Dataset}} & \multirow{2}{*}{\textbf{GT}}
& \multicolumn{3}{c@{}}{\textbf{Pred.}\,/\,\textbf{Match} nodes} \\
\cmidrule(l){3-5}
& & \textbf{Qwen3-8B} & \textbf{Qwen3-14B} & \textbf{GPT-4o} \\
\midrule
Sachs     & 11 & 11\,/\,6 & 11\,/\,8 & 12\,/\,11 \\
CP        & 11 & 10\,/\,1 & 7\,/\,1  & 15\,/\,4  \\
Coma      & 5  & 4\,/\,0  & 6\,/\,0  & 8\,/\,0   \\
Gait      & 8  & 16\,/\,1 & 31\,/\,1 & 15\,/\,1  \\
Alzheimer & 9  & 15\,/\,0 & 11\,/\,0 & 17\,/\,1  \\
Agro      & 6  & 19\,/\,0 & 8\,/\,0  & 12\,/\,0  \\
\bottomrule
\end{tabular}%
}
\caption{Data contamination testing. GT, Pred., and Match are the numbers of
ground-truth, LLM-predicted, and exactly matched node names.}
\label{tab:data_contamination}
\end{table}

\paragraph{Baselines.} 
We compare three types of baselines: (1) \textbf{Data-driven methods}: PC~\cite{spirtes2001causation}, GES~\cite{chickering2002optimal}, PC-ANM~\cite{peters2014causal}, DirectLiNGAM~\cite{shimizu2011directlingam}, NOTEARS~\cite{zheng2018dags}, DAG-GNN~\cite{yu2019dag} and DAGMA-L~\cite{bello2022dagma}. (2) \textbf{LLM-only methods}: GPT-4o~\cite{openai2024gpt4o}, Qwen3-8B and Qwen3-14B~\cite{yang2025qwen3}. (3) \textbf{Hybrid LLM-based methods}: LLM-greedy~\cite{long2023causal} and LLM-CD~\cite{du2025causal}. For the MIMIC dataset, additional baselines include FCI~\cite{spirtes2013causal}, GPT-3.5, GPT-4, GPT-4o-mini~\footnote{https://openai.com/api/},
LLM-BFS~\cite{jiralerspong2024efficient}, and ChatPC~\cite{cohrs2023large}. Appendix~\ref{appendix:G} provides more details.
\paragraph{Metrics.}
We evaluate graph recovery using standard metrics from prior causal discovery work~\cite{du2025causal}. We report F1 in the main text as a compact measure of edge-recovery accuracy. Precision, recall, SHD, NHD, and Ratio are provided in Appendix~\ref{metrics details}. For MIMIC-IV, following~\cite{du2025causal}, we report predictive metrics, including precision, recall, accuracy, and AUC. 
\paragraph{Implementation Details. }
LLM-CR pairs a GES backbone~\cite{chickering2002optimal} with GPT-4o.  A single schema is generated for each dataset, after which every pairwise decision is verified using deterministic decoding (\(temperature=0\)). The semantic branch obtains 10 independently sampled judgments for every ordered variable pair (\(temperature=0.1\)) and accepts a majority label when at least seven judgments agree. We use a fixed conflict-resolution configuration across all datasets:
\(\lambda_{\mathrm{BIC}}
,
\beta_{\mathrm{HSIC}}
,
\gamma_{\mathrm{LLM}}= (0.40,0.25,0.35)\), with a sparse prior \(\pi=0.1\) and a recovery threshold \(\tau_r=0.9\) selected using the Sachs dataset and evaluated on other datasets under fixed settings (Appendix~\ref{senstivity_analysis}). For the MIMIC-IV experiment, we follow the downstream prediction setting of ~\cite{du2025causal} and employ stratified five-fold cross-validation.

\subsection{RQ1: Data contamination}
Table~\ref{tab:data_contamination} evaluates the extent to which the benchmark variable names can be recovered from the LLMs' parametric knowledge. Except for the Sachs dataset, node-name recall is substantially lower across all benchmark datasets. The Sachs dataset shows higher recall across all evaluated models, suggesting its variable names are more likely to have appeared in the pre-training corpora aligned with~\cite{babakov2025scalability}. Moreover, the LLMs frequently generate far more variable names than appear in the ground-truth graphs while matching few or none of the benchmark variables, indicating that most predictions are unrelated or hallucinated rather than memorized benchmark nodes. Following the evaluation protocol of~\cite{babakov2025scalability}, we base our principal conclusions exclusively on the five lower-recall benchmarks and report Sachs only as an auxiliary result. We treat the remaining benchmarks as lower-risk, while acknowledging that the diagnostic cannot rule out prior exposure.

\subsection{RQ2: Main Results}
We first evaluate \textsc{LLM-CR} structural recovery performance against the baselines on all datasets except MIMIC-IV.  Table~\ref{tab:all-datasets-f1} reports compact F1 results for five datasets, but Appendix~\ref {detailed results} provides detailed results for all structural datasets, including Sachs. 
As shown in Table~\ref{tab:all-datasets-f1}, LLM-CR achieves the highest F1 on four of the five causal-structure datasets. It outperforms the strongest competing baseline by $22.3\%$ on Coma, $21.3\%$ on Agro, $20.2\%$ on Alzheimer, and $42.0\%$ on CP. On Gait, which comprises variables aggregated from multiple variables, GES remains the strongest method, achieving an F1 of $0.786$, while LLM-CR obtains $0.683$, which is $13.1\%$ lower than GES.
These results indicate that LLM-CR generally provides substantial improvements over the strongest baselines.
The results also expose a key limitation of purely data-driven methods: their structural recovery is sensitive to data-generating assumptions and distributional properties. Prompt-only LLMs benefit from variable semantics and metadata, with Qwen3-8B and Qwen3-14B with metadata achieving the competitive baseline NHD on the Coma dataset. LLM-CD integrates LLM knowledge into PC yet performs poorly, suggesting that early LLM-guided screening or PC-stage decisions may propagate errors on harder datasets. \textsc{LLM-CR} avoids this by applying LLM judgments only to disputed edges and weighting them by semantic-entropy uncertainty, preserving recall gains while reducing instability (See Appendix~\ref{sec: case study} for details). We report LLM calls, token usage, estimated API cost, and measured wall time for each dataset in Appendix~\ref{app:efficiency}.

\subsection{RQ3: Downstream Clinical Prediction}
\label{sec:rq2}

\definecolor{stdgray}{gray}{0.45}
\newcommand{\result}[2]{%
  \shortstack[c]{%
    #1\\[-1.5pt]
    \hspace*{+1.05em}{\scriptsize\itshape\color{stdgray} $\pm$\,#2}%
  }%
}

\begin{table}[t]
\centering
\small
\setlength{\tabcolsep}{1.5pt}
\renewcommand{\arraystretch}{1.05}
\resizebox{0.95\linewidth}{!}{%
\begin{tabular}{@{}lcccc@{}}
\toprule
\textbf{Model}
& \textbf{ACC$\uparrow$}
& \textbf{Recall$\uparrow$}
& \textbf{Prec.$\uparrow$}
& \textbf{AUC$\uparrow$} \\
\midrule

PC
& \result{0.673}{0.0056}
& \result{0.812}{0.0103}
& \result{0.637}{0.0025}
& \result{0.535}{0.0026} \\

FCI
& \result{0.558}{0.0041}
& \result{1.000}{0.0013}
& \result{0.527}{0.0007}
& \result{0.550}{0.0139} \\

\midrule

GPT-3.5
& \result{0.525}{0.0062}
& \result{1.000}{0.0015}
& \result{0.509}{0.0054}
& \result{0.551}{0.0192} \\

GPT-4
& \result{0.558}{0.0062}
& \result{1.000}{0.0019}
& \result{0.527}{0.0027}
& \result{0.532}{0.0147} \\

GPT-4o
& \result{0.557}{0.0081}
& \result{1.000}{0.0031}
& \result{0.527}{0.0104}
& \result{0.520}{0.0124} \\

GPT-4o-mini
& \result{0.558}{0.0118}
& \result{1.000}{0.0016}
& \result{0.527}{0.0159}
& \result{0.569}{0.0192} \\

\midrule

LLM-BFS
& \result{0.558}{0.0063}
& \result{1.000}{0.0072}
& \result{0.527}{0.0031}
& \result{0.604}{0.0186} \\

ChatPC
& \result{0.547}{0.0053}
& \result{0.981}{0.0162}
& \result{0.530}{0.0043}
& \result{0.613}{0.0181} \\

LLM-greedy
& \result{0.558}{0.0058}
& \result{1.000}{0.0113}
& \result{0.527}{0.0046}
& \result{0.608}{0.0158} \\

LLM-CD
& \result{0.573}{0.0022}
& \result{1.000}{0.0036}
& \result{0.538}{0.0011}
& \result{0.645}{0.0169} \\

\midrule

\textbf{LLM-CR}
& \result{0.569}{0.0131}
& \result{1.000}{0.0010}
& \result{0.539}{0.0076}
& \result{\textbf{0.784}}{0.0122} \\

\bottomrule
\end{tabular}
}


\caption{Five-fold out-of-fold classification performance on
MIMIC-IV. Values are reported as mean $\pm$ standard deviation
across folds. ACC denotes accuracy and Prec.\ denotes precision.}
\label{tab:mimic-results}
\end{table}

For the MIMIC dataset, we perform downstream prediction for lung cancer patients following~\cite{du2025causal}. The aim is to find all patients with lung cancer. LLM-CR identifies and evaluates the parent set for lung cancer prediction rather than causal correctness, achieving the highest AUC and second-highest precision among all methods, indicating better support for patient-level lung cancer risk stratification across thresholds. In Table~\ref{tab:mimic-results}, while PC achieves the highest precision, it has a lower AUC, indicating a limited ability to rank patients with lung cancer. These results suggest that the variables selected by LLM-CR provide useful predictive information for downstream lung cancer risk stratification.
\subsection{RQ4: Ablation Studies}








\begin{table}[t!]
\centering
\small
\setlength{\tabcolsep}{5pt}
\renewcommand{\arraystretch}{1.06}
\resizebox{\columnwidth}{!}{%
\begin{tabular}{lcccc}
\toprule
\multirow{2}{*}{\textbf{Method}}
& \multicolumn{2}{c}{\textbf{Alzheimer}}
& \multicolumn{2}{c}{\textbf{CP}} \\
\cmidrule(lr){2-3} \cmidrule(lr){4-5}
& \textbf{F1$\uparrow$}
& \textbf{NHD$\downarrow$}
& \textbf{F1$\uparrow$}
& \textbf{NHD$\downarrow$} \\
\midrule
Full
& \textbf{0.881} & \textbf{0.064} & \textbf{0.805} & \textbf{0.115} \\
\midrule
w/o Metadata
& 0.651 & 0.161 & 0.690 & 0.164 \\
w/o Schema
& 0.567 & 0.186 & 0.714 & 0.145 \\
w/o Semantic Entropy
& 0.741 & 0.116 & 0.783 & 0.136 \\
\midrule
Data-DAG
& 0.335 & 0.300 & 0.502 & 0.282 \\
\bottomrule
\end{tabular}
}
\caption{Ablation studies on the Alzheimer and CP datasets.}
\label{tab:ablation}
\end{table}

\begin{figure}[t]
  \centering
  \includegraphics[width=\columnwidth]{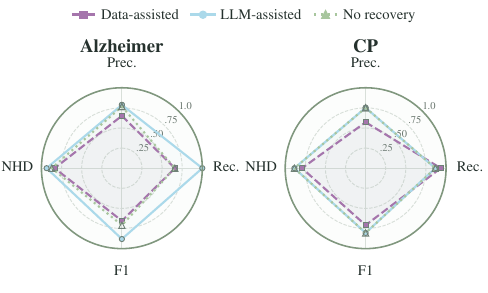}
  \caption{Recovery-stage analysis on the Alzheimer and CP datasets. Precision, recall, and F1 are shown directly. NHD is inverted so that higher values indicate better performance.} 
  \label{fig:line-chart}
\end{figure}
\begin{figure}[t]
  \centering
  \includegraphics[width=0.95\columnwidth]{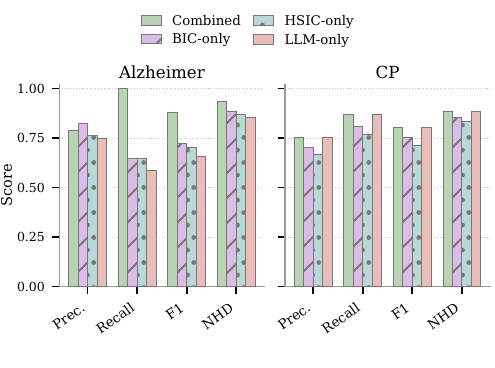}
  \caption{Score-component analysis on the Alzheimer and CP datasets.
 Precision, recall, and F1 are shown directly. NHD is inverted so that higher values indicate better performance. }
  \label{fig:bar-chart}
\end{figure}
\label{sec:rq3}
We conduct an ablation study by removing four key components of LLM-CR: (1) \textit{w/o Metadata}, which removes variable descriptions from the LLM prompt; (2) \textit{w/o Schema}, which removes schema-guided constraints during reasoning; (3) \textit{w/o Semantic Entropy}, which disables uncertainty-aware calibration of the LLM prior; and (4) \textit{Data-DAG}, which relies solely on the data-driven graph without LLM guidance. As shown in Table~\ref{tab:ablation}, each component contributes to LLM-CR's overall performance, demonstrating that the framework benefits from the complementary integration of structured knowledge and statistical evidence. In both datasets, removing metadata consistently degrades structural recovery, indicating that semantically rich variable descriptions provide essential context for reliable causal reasoning. Eliminating schema-guided constraints substantially reduces F1 across both datasets, highlighting the role of structured reasoning in preventing inconsistent or incomplete causal hypotheses. Disabling semantic-entropy calibration degrades overall structural accuracy on both datasets, suggesting that uncalibrated LLM predictions may introduce unreliable causal hypotheses and affect graph structural quality. Data-DAG's lower performance in these experiments suggests that statistical evidence alone was insufficient to achieve the structural recovery obtained by the full framework. 

\subsection{RQ5: Scoring and Recovery Components}
\label{sec:rq4}
Figure~\ref{fig:line-chart} compares LLM-assisted recovery, data-assisted recovery, and the no-recovery variant. LLM-assisted recovery considers omitted edges with strong LLM support, whereas data-assisted recovery revisits omitted edges proposed by the data-driven branch. In data-assisted recovery, all missing edges are revisited, and their scores are recomputed, improving recall in CP data, but overall structural accuracy drops because many false edges are also recovered. For CP, LLM-assisted recovery did not identify any additional candidate whose score exceeded the recovery threshold. In the Alzheimer dataset, recall is lower for the data-assisted and no-recovery stages, but it improves in the LLM-assisted recovery stage, which also balances overall structural accuracy. Figure~\ref{fig:bar-chart} highlights the importance of different components of the overall structural accuracy score. Each score component tends to add more false positives when used separately to resolve the conflict. However, when combined with their weights, they contribute more to true edge recovery across both datasets.

\section{Conclusion}
This work presented an uncertainty-aware framework that resolves the conflict between LLMs and data-driven causal discovery. The framework treats schema-guided LLMs as semantic causal knowledge sources without making them final decision-makers, and combines them with data-driven causal discovery using a score that combines data fit, dependence evidence, and uncertainty-aware LLM prior evidence. The results show that schema guidance improves the LLM-driven causal reasoning while conflict resolution reduces unsupported edges. Across different datasets, our framework achieves a strong balance between recall, precision, and structural accuracy. Overall, this work demonstrates LLM-assisted causal discovery through verification and evidence-based graph refinement.
\section{Limitations}
\textsc{LLM-CR} is designed for settings where variable names and metadata provide meaningful semantic context. When such information is limited, the schema-guided LLM branch may contribute less additional evidence, and the framework naturally relies more on the data-driven component. This does not affect the general conflict-resolution design, but it may reduce the benefit of incorporating LLM-derived priors.
The current score combines BIC, HSIC, and an uncertainty-calibrated LLM prior; future work can replace or extend these components for different data regimes. While \textsc{LLM-CR} enforces DAG validity, explicit treatment of latent confounding remains an important future direction.
\section{Ethical Considerations}
This work presents a research framework for causal discovery, not an automated decision-making system. In clinical settings, interpret any learned graph with domain-expert review, and users should consult qualified healthcare professionals for medical concerns.
\section*{Acknowledgments}
This research was supported by the MRFF-funded Motion Connect project (MRFFRD11000028). The first author acknowledges financial support from the Griffith University Postgraduate Research Scholarship and the Griffith University International Postgraduate Research Scholarship.

\bibliography{custom}

@article{glymour2019review,
  title={Review of causal discovery methods based on graphical models},
  author={Glymour, Clark and Zhang, Kun and Spirtes, Peter},
  journal={Frontiers in genetics},
  volume={10},
  pages={524},
  year={2019},
  publisher={Frontiers Media SA},
  doi = {10.3389/fgene.2019.00524}
}

@article{nichol2010challenging,
  title={Challenging issues in randomised controlled trials},
  author={Nichol, A. D. and
            Bailey, M. and
            Cooper, D. J. and
            {POLAR Study Investigators} and
            {EPO-TBI Study Investigators} and
            Rosenfeld, Jeffrey},
  journal={Injury},
  volume={41},
  pages={S20--S23},
  year={2010},
  publisher={Elsevier},
  doi = {10.1016/j.injury.2010.03.033},
}

@article{zanga2022survey,
  title={A survey on causal discovery: Theory and practice},
  author={Zanga, Alessio and Ozkirimli, Elif and Stella, Fabio},
  journal={International Journal of Approximate Reasoning},
  volume={151},
  pages={101--129},
  year={2022},
  publisher={Elsevier},
  doi = {10.1016/j.ijar.2022.09.004}
}

@book{spirtes2001causation,
  title = {Causation, Prediction, and Search},
  author = {Spirtes, Peter and
            Glymour, Clark and
            Scheines, Richard},
  edition = {2},
  publisher = {MIT Press},
  address = {Cambridge, MA},
  year = {2001},
  isbn = {9780262194402},
  url = {https://mitpress.mit.edu/9780262527927/causation-prediction-and-search/}
}

@article{spirtes2013causal,
  title={Causal inference in the presence of latent variables and selection bias},
  author={Spirtes, Peter L and Meek, Christopher and Richardson, Thomas S},
  journal={arXiv preprint arXiv:1302.4983},
  year={2013},
  url={https://arxiv.org/abs/1302.4983}
}

@article{colombo2012learning,
  title={Learning high-dimensional directed acyclic graphs with latent and selection variables},
  author={Colombo, Diego and Maathuis, Marloes H and Kalisch, Markus and Richardson, Thomas S},
  journal={The Annals of Statistics},
  volume = {40},
  number = {1},
  pages={294--321},
  year={2012},
  publisher={JSTOR},
  doi = {10.1214/11-AOS940}
}

@article{squires2023causal,
  title={Causal structure learning: A combinatorial perspective},
  author={Squires, Chandler and Uhler, Caroline},
  journal={Foundations of Computational Mathematics},
  volume={23},
  number={5},
  pages={1781--1815},
  year={2023},
  publisher={Springer},
  doi = {10.1007/s10208-022-09581-9}
}

@article{chickering2002optimal,
  title={Optimal structure identification with greedy search},
  author={Chickering, David Maxwell},
  journal={Journal of machine learning research},
  volume={3},
  number={Nov},
  pages={507--554},
  year={2002},
  url = {https://jmlr.org/papers/v3/chickering02b.html}
}

@article{ramsey2017million,
  title={A million variables and more: the fast greedy equivalence search algorithm for learning high-dimensional graphical causal models, with an application to functional magnetic resonance images},
  author={Ramsey, Joseph and Glymour, Madelyn and Sanchez-Romero, Ruben and Glymour, Clark},
  journal={International journal of data science and analytics},
  volume={3},
  number={2},
  pages={121--129},
  year={2017},
  publisher={Springer},
  doi = {10.1007/s41060-016-0032-z}
}

@inproceedings{zhou2022survey,
  title = {A Survey on Causal Discovery},
  author = {Zhou, Wenxiu and Chen, Qingcai},
  booktitle = {Knowledge Graph and Semantic Computing: Knowledge Graph Empowers the Digital Economy},
  series = {Communications in Computer and Information Science},
  volume = {1669},
  pages = {123--135},
  year = {2022},
  publisher = {Springer},
  doi = {10.1007/978-981-19-7596-7_10},
  url = {https://doi.org/10.1007/978-981-19-7596-7_10}
}

@article{zheng2018dags,
  title = {{DAGs} with {NO TEARS}: Continuous Optimization for Structure Learning},
  author = {Zheng, Xun and
            Aragam, Bryon and
            Ravikumar, Pradeep K. and
            Xing, Eric P.},
  booktitle = {Advances in Neural Information Processing Systems},
  volume = {31},
  pages = {9492--9503},
  year = {2018},
  url = {https://proceedings.neurips.cc/paper/2018/hash/e347c51419ffb23ca3fd5050202f9c3d-Abstract.html}
}

@article{wang2024survey,
  title={A survey of causal discovery based on functional causal model},
  author={Wang, Lei and Huang, Shanshan and Wang, Shu and Liao, Jun and Li, Tingpeng and Liu, Li},
  journal={Engineering Applications of Artificial Intelligence},
  volume={133},
  pages={108258},
  year={2024},
  publisher={Elsevier},
  doi = {10.1016/j.engappai.2024.108258}
}

@inproceedings{yu2019dag,
  title = {{DAG-GNN}: {DAG} Structure Learning with Graph Neural Networks},
  author = {Yu, Yue and
            Chen, Jie and
            Gao, Tian and
            Yu, Mo},
  booktitle = {Proceedings of the 36th International Conference on Machine Learning},
  series = {Proceedings of Machine Learning Research},
  volume = {97},
  pages = {7154--7163},
  year = {2019},
  editor = {Chaudhuri, Kamalika and Salakhutdinov, Ruslan},
  publisher = {PMLR},
  url = {https://proceedings.mlr.press/v97/yu19a.html}
}

@article{steeger2021common,
  title = {Common Methodological Problems in Randomized Controlled Trials of Preventive Interventions},
  author = {Steeger, Christine M. and
            Buckley, Pamela R. and
            Pampel, Fred C. and
            Gust, Charleen J. and
            Hill, Karl G.},
  journal = {Prevention Science},
  volume = {22},
  number = {8},
  pages = {1159--1172},
  year = {2021},
  doi = {10.1007/s11121-021-01263-2},
  url = {https://doi.org/10.1007/s11121-021-01263-2}
}

@article{peters2014causal,
  title = {Causal Discovery with Continuous Additive Noise Models},
  author = {Peters, Jonas and
            Mooij, Joris M. and
            Janzing, Dominik and
            Sch{\"o}lkopf, Bernhard},
  journal = {Journal of Machine Learning Research},
  volume = {15},
  number = {58},
  pages = {2009--2053},
  year = {2014},
  url = {https://jmlr.org/papers/v15/peters14a.html}
}

@misc{peng2026causcientist,
  title = {{CauScientist}: Teaching {LLMs} to Respect Data for Causal Discovery},
  author = {Peng, Bo and
            Chen, Sirui and
            Xu, Lei and
            Lu, Chaochao},
  year = {2026},
  eprint = {2601.13614},
  archivePrefix = {arXiv},
  primaryClass = {cs.CL},
  doi = {10.48550/arXiv.2601.13614},
  url = {https://arxiv.org/abs/2601.13614}
}

@article{
kiciman2024causal,
title={Causal Reasoning and Large Language Models: Opening a New Frontier for Causality},
author={Emre Kiciman and Robert Ness and Amit Sharma and Chenhao Tan},
journal={Transactions on Machine Learning Research},
issn={2835-8856},
year={2024},
url={https://openreview.net/forum?id=mqoxLkX210},
note={Featured Certification}
}

@inproceedings{du2025causal,
  title={Causal discovery through synergizing large language model and data-driven reasoning},
  author={Du, Huaming and Zheng, Yujia and Jing, Baoyu and Zhao, Yu and Kou, Gang and Liu, Guisong and Gu, Tao and Li, Weimin and Yang, Carl},
  booktitle={Proceedings of the 31st ACM SIGKDD Conference on Knowledge Discovery and Data Mining V. 2},
  pages={543--554},
  year={2025},
   doi = {10.1145/3711896.3736874}
}

@inproceedings{
long2022can,
title={Can Large Language Models Build Causal Graphs?},
author={Stephanie Long and Tibor Schuster and Alexandre Pich{\'e}},
booktitle={NeurIPS 2022 Workshop on Causality for Real-world Impact},
year={2022},
url={https://openreview.net/forum?id=LQQoJGw8JD1}
}

@article{ban2025llm,
  title={Llm-driven causal discovery via harmonized prior},
  author={Ban, Taiyu and Chen, Lyuzhou and Lyu, Derui and Wang, Xiangyu and Zhu, Qinrui and Chen, Huanhuan},
  journal={IEEE Transactions on Knowledge and Data Engineering},
  volume={37},
  number={4},
  pages={1943--1960},
  year={2025},
  publisher={IEEE},
  doi = {10.1109/TKDE.2025.3528461}
}

@article{wang2025large,
  title={Large-scale hierarchical causal discovery via weak prior knowledge},
  author={Wang, Xiangyu and Ban, Taiyu and Chen, Lyuzhou and Lyu, Derui and Zhu, Qinrui and Chen, Huanhuan},
  journal={IEEE Transactions on Knowledge and Data Engineering},
  year={2025},
  publisher={IEEE},
  doi = {10.1109/TKDE.2025.3537832}
}

@inproceedings{abdulaal2024causal,
  title={Causal modelling agents: Causal graph discovery through synergising metadata-and data-driven reasoning},
  author={Abdulaal, Ahmed and Hadjivasiliou, Adamos and Montana-Brown, Nina and He, Tiantian and Ijishakin, Ayodeji and Drobnjak, Ivana and Castro, Daniel and Alexander, Daniel},
  booktitle={International Conference on Learning Representations},
  year={2024},
  url = {https://openreview.net/forum?id=pAoqRlTBtY}
}

@article{
zevcevic2023causal,
title={Causal Parrots: Large Language Models May Talk Causality But Are Not Causal},
author={Matej Ze{\v{c}}evi{\'c} and Moritz Willig and Devendra Singh Dhami and Kristian Kersting},
journal={Transactions on Machine Learning Research},
issn={2835-8856},
year={2023},
url={https://openreview.net/forum?id=tv46tCzs83},
note={}
}

@inproceedings{
long2023causal,
title={Causal Discovery with Language Models as Imperfect Experts},
author={Stephanie Long and Alexandre Pich{\'e} and Valentina Zantedeschi and Tibor Schuster and Alexandre Drouin},
booktitle={ICML 2023 Workshop on Structured Probabilistic Inference {\&} Generative Modeling},
year={2023},
}

@article{khatibi2024alcm,
  title={Alcm: Autonomous llm-augmented causal discovery framework},
  author={Khatibi, Elahe and Abbasian, Mahyar and Yang, Zhongqi and Azimi, Iman and Rahmani, Amir M},
  journal={arXiv preprint arXiv:2405.01744},
  year={2024},
  doi = {10.48550/arXiv.2405.01744}
}

@article{darvariu2024large,
  title={Large language models are effective priors for causal graph discovery},
  author={Darvariu, Victor-Alexandru and Hailes, Stephen and Musolesi, Mirco},
  journal={arXiv preprint arXiv:2405.13551},
  year={2024},
  doi = {10.48550/arXiv.2405.13551}
}

@inproceedings{jin2023cladder,
  title = {{CLadder}: Assessing Causal Reasoning in Language Models},
  author = {Jin, Zhijing and
            Chen, Yuen and
            Leeb, Felix and
            Gresele, Luigi and
            Kamal, Ojasv and
            Lyu, Zhiheng and
            Blin, Kevin and
            Adauto, Fernando Gonzalez and
            Kleiman-Weiner, Max and
            Sachan, Mrinmaya and
            Sch{\"o}lkopf, Bernhard},
  booktitle = {Advances in Neural Information Processing Systems},
  volume = {36},
  pages = {31038--31065},
  year = {2023},
  doi = {10.52202/075280-1353}
}

@inproceedings{
cohrs2023large,
title={Large Language Models for Constrained-Based Causal Discovery},
author={Kai-Hendrik Cohrs and Emiliano Diaz and Vasileios Sitokonstantinou and Gherardo Varando and Gustau Camps-Valls},
booktitle={AAAI 2024 Workshop on ''Are Large Language Models Simply Causal Parrots?''},
year={2023},
url={https://openreview.net/forum?id=NEAoZRWHPN}
}

@inproceedings{vashishtha2025causal,
  title = {Causal Order: The Key to Leveraging Imperfect Experts in Causal Inference},
  author = {Vashishtha, Aniket and
            Reddy, Abbavaram Gowtham and
            Kumar, Abhinav and
            Bachu, Saketh and
            Balasubramanian, Vineeth N. and
            Sharma, Amit},
  booktitle = {The Thirteenth International Conference on Learning Representations},
  year = {2025},
  url = {https://openreview.net/forum?id=9juyeCqL0u}
}

@article{ban2025integrating,
  title={Integrating large language model for improved causal discovery},
  author={Ban, Taiyu and Chen, Lyuzhou and Lyu, Derui and Wang, Xiangyu and Zhu, Qinrui and Tu, Qiang and Chen, Huanhuan},
  journal={IEEE Transactions on Artificial Intelligence},
  volume={6},
  number={11},
  pages={3030--3042},
  year={2025},
  publisher={IEEE},
  doi = {10.1109/TAI.2025.3560927}
}

@inproceedings{wan2025large,
  title = {Large Language Models for Causal Discovery: Current Landscape and Future Directions},
  author = {Wan, Guangya and
            Lu, Yunsheng and
            Wu, Yuqi and
            Hu, Mengxuan and
            Li, Sheng},
  booktitle = {Proceedings of the Thirty-Fourth International Joint Conference on Artificial Intelligence},
  pages = {10687--10695},
  year = {2025},
  doi = {10.24963/ijcai.2025/1186},
  url = {https://doi.org/10.24963/ijcai.2025/1186}
}

@inproceedings{xiong2024can,
  title={Can llms express their uncertainty? an empirical evaluation of confidence elicitation in llms},
  author={Xiong, Miao and Hu, Zhiyuan and Lu, Xinyang and Li, Yifei and Fu, Jie and He, Junxian and Hooi, Bryan},
  booktitle={International Conference on Learning Representations},
  year={2024},
   url = {https://proceedings.iclr.cc/paper_files/paper/2024/hash/6733cf15e10e2cd1d59af033c3bb8507-Abstract-Conference.html}
}

@article{farquhar2024detecting,
  title={Detecting hallucinations in large language models using semantic entropy},
  author={Farquhar, Sebastian and Kossen, Jannik and Kuhn, Lorenz and Gal, Yarin},
  journal={Nature},
  volume={630},
  number={8017},
  pages={625--630},
  year={2024},
  publisher={Nature Publishing Group UK London},
   doi = {10.1038/s41586-024-07421-0}
}

@article{sachs2005causal,
  title={Causal protein-signaling networks derived from multiparameter single-cell data},
  author={Sachs, Karen and Perez, Omar and Pe'er, Dana and Lauffenburger, Douglas A and Nolan, Garry P},
  journal={Science},
  volume={308},
  number={5721},
  pages={523--529},
  year={2005},
  publisher={American Association for the Advancement of Science},
  doi = {10.1126/science.1105809}
}

@article{zhang2025siren,
  title={ Siren’s Song in the AI Ocean: A Survey on Hallucination in Large Language Models},
  author = {Zhang, Yue and
            Li, Yafu and
            Cui, Leyang and
            Cai, Deng and
            Liu, Lemao and
            Fu, Tingchen and
            Huang, Xinting and
            Zhao, Enbo and
            Zhang, Yu and
            Chen, Yulong and
            Wang, Longyue and
            Luu, Anh Tuan and
            Bi, Wei and
            Shi, Freda and
            Shi, Shuming},
  journal={Computational Linguistics},
  volume={51},
  number={4},
  pages={1373--1418},
  year={2025},
  publisher={MIT Press 255 Main Street, 9th Floor, Cambridge, Massachusetts 02142, USA~…},
  doi = {10.1162/coli.a.16}
}

@article{johnson2023mimic,
  title = {{MIMIC-IV}, a Freely Accessible Electronic Health Record Dataset},
  author = {Johnson, Alistair E. W. and
            Bulgarelli, Lucas and
            Shen, Lu and
            Gayles, Alvin and
            Shammout, Ayad and
            Horng, Steven and
            Pollard, Tom J. and
            Hao, Sicheng and
            Moody, Benjamin and
            Gow, Brian and
            Lehman, Li-Wei H. and
            Celi, Leo A. and
            Mark, Roger G.},
  journal = {Scientific Data},
  volume = {10},
  number = {1},
  pages = {1},
  year = {2023},
  doi = {10.1038/s41597-022-01899-x},
  url = {https://doi.org/10.1038/s41597-022-01899-x}
}

@article{gill2023causal,
  title={Causal modelling demonstrates metabolic power is largely affected by gait kinematics and motor control in children with cerebral palsy},
  author={Gill, Pavreet K and Steele, Katherine M and Donelan, J Maxwell and Schwartz, Michael H},
  journal={PLoS One},
  volume={18},
  number={5},
  pages={e0285667},
  year={2023},
  publisher={Public Library of Science San Francisco, CA USA},
  doi = {10.1371/journal.pone.0285667}
}

@article{gretton2007kernel,
  title={A kernel statistical test of independence},
  author={Gretton, Arthur and Fukumizu, Kenji and Teo, Choon and Song, Le and Sch{\"o}lkopf, Bernhard and Smola, Alex},
  journal={Advances in neural information processing systems},
  volume={20},
  pages = {585--592},
  year={2007},
  url = {https://proceedings.neurips.cc/paper_files/paper/2007/hash/d5cfead94f5350c12c322b5b664544c1-Abstract.html} 
}

@article{hoyer2008nonlinear,
  title={Nonlinear causal discovery with additive noise models},
  author={Hoyer, Patrik and Janzing, Dominik and Mooij, Joris M and Peters, Jonas and Sch{\"o}lkopf, Bernhard},
  journal={Advances in neural information processing systems},
  volume = {21},
  pages = {689--696},
  year = {2008},
  url = {https://proceedings.neurips.cc/paper_files/paper/2008/hash/f7664060cc52bc6f3d620bcedc94a4b6-Abstract.html}
}

@article{shimizu2011directlingam,
  title = {{DirectLiNGAM}: A Direct Method for Learning a Linear Non-Gaussian Structural Equation Model},
  author = {Shimizu, Shohei and
            Inazumi, Takanori and
            Sogawa, Yasuhiro and
            Hyv{\"a}rinen, Aapo and
            Kawahara, Yoshinobu and
            Washio, Takashi and
            Hoyer, Patrik O. and
            Bollen, Kenneth},
  journal = {Journal of Machine Learning Research},
  volume = {12},
  number = {33},
  pages = {1225--1248},
  year = {2011},
  url = {https://www.jmlr.org/papers/v12/shimizu11a.html}
}

@article{bello2022dagma,
  title={{DAGMA}: Learning {DAGs} via {M}-matrices and a Log-determinant Acyclicity Characterization},
  author={Bello, Kevin and Aragam, Bryon and Ravikumar, Pradeep},
  journal={Advances in Neural Information Processing Systems},
  volume={35},
  pages={8226--8239},
  year={2022},
  doi = {10.52202/068431-0598}
}

@article{openai2024gpt4o,
  title = {{GPT-4o} System Card},
  author = {{OpenAI}},
  journal = {arXiv preprint arXiv:2410.21276},
  year = {2024},
  doi = {10.48550/arXiv.2410.21276},
  url = {https://arxiv.org/abs/2410.21276}
}

@article{yang2025qwen3,
  title = {{Qwen3} Technical Report},
  author = {{Qwen Team}},
  year = {2025},
  eprint = {2505.09388},
  archivePrefix = {arXiv},
  primaryClass = {cs.CL},
  doi = {10.48550/arXiv.2505.09388},
  url = {https://arxiv.org/abs/2505.09388}
}

@article{scutari2010learning,
  title={Learning Bayesian networks with the bnlearn R package},
  author={Scutari, Marco},
  journal={Journal of statistical software},
  volume={35},
  pages={1--22},
  year={2010},
  doi = {10.18637/jss.v035.i03}
}

@inproceedings{mooij2020constraint,
  title={Constraint-based causal discovery using partial ancestral graphs in the presence of cycles},
  author={Mooij, Joris M and Claassen, Tom},
  booktitle={Conference on Uncertainty in Artificial Intelligence},
  pages={1159--1168},
  year={2020},
  publisher = {PMLR},
  url = {https://proceedings.mlr.press/v124/m-mooij20a.html}
}

@article{chi2024unveiling,
  title={Unveiling causal reasoning in large language models: Reality or mirage?},
  author={Chi, Haoang and Li, He and Yang, Wenjing and Liu, Feng and Lan, Long and Ren, Xiaoguang and Liu, Tongliang and Han, Bo},
  journal={Advances in Neural Information Processing Systems},
  volume={37},
  pages={96640--96670},
  year={2024},
  doi = {10.52202/079017-3064}
}

@inproceedings{
jiralerspong2024efficient,
title={Efficient Causal Graph Discovery Using Large Language Models},
author={Thomas Jiralerspong and Xiaoyin Chen and Yash More and Vedant Shah and Yoshua Bengio},
booktitle={ICLR 2024 Workshop: How Far Are We From AGI},
year={2024},
url={https://openreview.net/forum?id=5RBUTx75yr}
}

@inproceedings{nazaret2024extremely,
  title = {Extremely Greedy Equivalence Search},
  author = {Nazaret, Achille and Blei, David},
  booktitle = {Proceedings of the Fortieth Conference on Uncertainty in Artificial Intelligence},
  series = {Proceedings of Machine Learning Research},
  volume = {244},
  pages = {2716--2745},
  year = {2024},
  editor = {Kiyavash, Negar and Mooij, Joris M.},
  publisher = {PMLR},
  url = {https://proceedings.mlr.press/v244/nazaret24a.html}
}

@inproceedings{babakov2025scalability,
  title={Scalability of {B}ayesian Network structure elicitation with Large Language Models: a novel methodology and comparative analysis},
  author={Babakov, Nikolay and Reiter, Ehud and Diz, Alberto Bugar{\'\i}n},
  booktitle={Proceedings of the 31st International Conference on Computational Linguistics},
  pages={10685--10711},
  year={2025},
  url = "https://aclanthology.org/2025.coling-main.713/"
}

@techreport{cooper1984nestor,
  title = {{NESTOR}: A Computer-Based Medical Diagnostic Aid That Integrates Causal and Probabilistic Knowledge},
  author = {Cooper, Gregory Floyd},
  institution = {Stanford University, Department of Computer Science},
  number = {STAN-CS-84-1031},
  year = {1984},
  address = {Stanford, California, USA},
  note = {Also HPP-84-48; PhD dissertation}
}

@article{nagesh2026evaluating,
  title={Evaluating Causal Discovery Algorithms for Path-Specific Fairness and Utility in Healthcare},
  author={Nagesh, Nitish and Khatibi, Elahe and Hughes, Thomas and Bagheri, Mahdi and Gajane, Pratik and Rahmani, Amir M},
  journal={arXiv preprint arXiv:2603.15926v1},
  year={2026},
  eprint = {2603.15926},
  archivePrefix = {arXiv},
  url = {https://arxiv.org/html/2603.15926v1}
}

@article{steele2022causal,
  title={Causal effects of motor control on gait kinematics after orthopedic surgery in cerebral palsy: a machine-learning approach},
  author={Steele, Katherine M and Schwartz, Michael H},
  journal={Frontiers in Human Neuroscience},
  volume={16},
  pages={846205},
  year={2022},
  publisher={Frontiers Media SA},
  doi = {10.3389/fnhum.2022.846205}
}

@article{baudrit2022decision,
  title={Decision support tool for the agri-food sector using data annotated by ontology and Bayesian Network: A proof of concept applied to milk microfiltration},
  author={Baudrit, Cedric and Buche, Patrice and Leconte, Nadine and Fernandez, Christophe and Belna, Ma{\"e}llis and G{\'e}san-Guiziou, Genevi{\`e}ve},
  journal={International Journal of Agricultural and Environmental Information Systems (IJAEIS)},
  volume={13},
  number={1},
  pages={1--22},
  year={2022},
  publisher={IGI Global Scientific Publishing},
  doi = {10.4018/IJAEIS.309136},
}

@inproceedings{
he2021deberta,
title={{\{}DEBERTA{\}}: {\{}DECODING{\}}-{\{}ENHANCED{\}} {\{}BERT{\}} {\{}WITH{\}} {\{}DISENTANGLED{\}} {\{}ATTENTION{\}}},
author={Pengcheng He and Xiaodong Liu and Jianfeng Gao and Weizhu Chen},
booktitle={International Conference on Learning Representations},
year={2021},
url={https://openreview.net/forum?id=XPZIaotutsD}
}

@article{zhang2018large,
  title={Large-scale kernel methods for independence testing},
  author={Zhang, Qinyi and Filippi, Sarah and Gretton, Arthur and Sejdinovic, Dino},
  journal={Statistics and Computing},
  volume={28},
  number={1},
  pages={113--130},
  year={2018},
  publisher={Springer},
  doi = {10.1007/s11222-016-9721-7}
}

\newpage
\appendix

\section{Detailed Results}
\begin{table*}[t!]
\centering
\small
\setlength{\tabcolsep}{4pt}

\resizebox{\textwidth}{!}{%
\begin{tabular}{lcccccc}
\toprule
\textbf{Method} & \textbf{Prec.$\uparrow$} & \textbf{Rec.$\uparrow$} & \textbf{F1$\uparrow$} & \textbf{SHD$\downarrow$} & \textbf{NHD$\downarrow$} & \textbf{Ratio$\downarrow$} \\
\midrule
PC & 0.298 $\pm$ 0.070 & 0.133 $\pm$ 0.053 & 0.183 $\pm$ 0.064 & 35.000 $\pm$ 1.225 & 0.318 $\pm$ 0.011 & 1.000 $\pm$ 0.035 \\
GES & 0.487 $\pm$ 0.122 & 0.520 $\pm$ 0.135 & 0.502 $\pm$ 0.126 & 31.000 $\pm$ 8.396 & 0.282 $\pm$ 0.076 & 0.886 $\pm$ 0.240 \\
PC-ANM & 0.360 $\pm$ 0.068 & 0.173 $\pm$ 0.055 & 0.233 $\pm$ 0.064 & 33.800 $\pm$ 1.483 & 0.307 $\pm$ 0.013 & 0.966 $\pm$ 0.042 \\
Direct-LiNGAM & 0.291 $\pm$ 0.085 & 0.393 $\pm$ 0.086 & 0.334 $\pm$ 0.087 & 47.600 $\pm$ 8.142 & 0.433 $\pm$ 0.074 & 1.360 $\pm$ 0.233 \\
NOTEARS & 0.213 $\pm$ 0.197 & 0.040 $\pm$ 0.037 & 0.067 $\pm$ 0.061 & 33.000 $\pm$ 1.871 & 0.300 $\pm$ 0.017 & 0.943 $\pm$ 0.053 \\
DAG-GNN & 0.000 $\pm$ 0.000 & 0.000 $\pm$ 0.000 & 0.000 $\pm$ 0.000 & 30.000 $\pm$ 0.000 & 0.273 $\pm$ 0.000 & 0.857 $\pm$ 0.000 \\
DAGMA-L & 0.461 $\pm$ 0.076 & 0.113 $\pm$ 0.018 & 0.182 $\pm$ 0.029 & 30.600 $\pm$ 1.140 & 0.278 $\pm$ 0.010 & 0.874 $\pm$ 0.033 \\
\midrule
GPT-4o & 0.691 $\pm$ 0.081 & 0.253 $\pm$ 0.030 & 0.371 $\pm$ 0.044 & 25.800 $\pm$ 1.789 & 0.235 $\pm$ 0.016 & 0.737 $\pm$ 0.051 \\
\quad + metadata & 0.651 $\pm$ 0.062 & 0.460 $\pm$ 0.043 & 0.539 $\pm$ 0.051 & 23.600 $\pm$ 2.608 & 0.215 $\pm$ 0.024 & 0.674 $\pm$ 0.075 \\
Qwen3-8B & 0.514 $\pm$ 0.000 & \secondg{0.633 $\pm$ 0.000} & \secondg{0.567 $\pm$ 0.000} & 29.000 $\pm$ 0.000 & 0.264 $\pm$ 0.000 & 0.829 $\pm$ 0.000 \\
\quad + metadata & 0.529 $\pm$ 0.000 & 0.600 $\pm$ 0.000 & 0.563 $\pm$ 0.000 & 28.000 $\pm$ 0.000 & 0.255 $\pm$ 0.000 & 0.800 $\pm$ 0.000 \\
Qwen3-14B & \secondg{0.834 $\pm$ 0.110} & 0.300 $\pm$ 0.047 & 0.437 $\pm$ 0.044 & 23.000 $\pm$ 1.225 & 0.209 $\pm$ 0.011 & 0.657 $\pm$ 0.035 \\
\quad + metadata & \best{0.876 $\pm$ 0.045} & 0.287 $\pm$ 0.051 & 0.430 $\pm$ 0.060 & \secondg{22.600 $\pm$ 1.673} & \secondg{0.205 $\pm$ 0.015} & \secondg{0.646 $\pm$ 0.048} \\
\midrule
LLM-greedy & 0.200 $\pm$ 0.025 & 0.147 $\pm$ 0.018 & 0.169 $\pm$ 0.021 & 43.200 $\pm$ 1.095 & 0.393 $\pm$ 0.010 & 1.234 $\pm$ 0.031 \\
LLM-CD & 0.610 $\pm$ 0.052 & 0.340 $\pm$ 0.015 & 0.436 $\pm$ 0.022 & 26.400 $\pm$ 1.517 & 0.240 $\pm$ 0.014 & 0.754 $\pm$ 0.043 \\
\midrule
LLM-CR & 0.752 $\pm$ 0.012 & \best{0.867 $\pm$ 0.000} & \best{0.805 $\pm$ 0.007} & \best{12.600 $\pm$ 0.548} & \best{0.115 $\pm$ 0.005} & \best{0.360 $\pm$ 0.016} \\
\bottomrule
\end{tabular}
}
\caption{Five-run causal discovery performance on the CP dataset.}
\label{tab:cp-results}
\end{table*}

\begin{table*}[t!]
\centering
\small
\setlength{\tabcolsep}{4pt}

\resizebox{\textwidth}{!}{%
\begin{tabular}{lcccccc}
\toprule
\textbf{Method} & \textbf{Prec.$\uparrow$} & \textbf{Rec.$\uparrow$} & \textbf{F1$\uparrow$} & \textbf{SHD$\downarrow$} & \textbf{NHD$\downarrow$} & \textbf{Ratio$\downarrow$} \\
\midrule
PC & 0.120 $\pm$ 0.164 & 0.071 $\pm$ 0.097 & 0.089 $\pm$ 0.122 & 24.000 $\pm$ 2.828 & 0.333 $\pm$ 0.039 & 1.000 $\pm$ 0.118 \\
GES & 0.355 $\pm$ 0.098 & 0.318 $\pm$ 0.067 & 0.335 $\pm$ 0.081 & 21.600 $\pm$ 3.209 & 0.300 $\pm$ 0.045 & 0.900 $\pm$ 0.134 \\
PC-ANM & 0.115 $\pm$ 0.157 & 0.071 $\pm$ 0.097 & 0.087 $\pm$ 0.120 & 24.200 $\pm$ 2.588 & 0.336 $\pm$ 0.036 & 1.008 $\pm$ 0.108 \\
Direct-LiNGAM & 0.249 $\pm$ 0.024 & 0.294 $\pm$ 0.042 & 0.270 $\pm$ 0.031 & 27.000 $\pm$ 0.707 & 0.375 $\pm$ 0.010 & 1.125 $\pm$ 0.029 \\
NOTEARS & 0.200 $\pm$ 0.000 & 0.059 $\pm$ 0.000 & 0.091 $\pm$ 0.000 & 20.000 $\pm$ 0.000 & 0.278 $\pm$ 0.000 & 0.833 $\pm$ 0.000 \\
DAG-GNN & 0.163 $\pm$ 0.045 & 0.082 $\pm$ 0.032 & 0.109 $\pm$ 0.038 & 22.600 $\pm$ 0.548 & 0.314 $\pm$ 0.008 & 0.942 $\pm$ 0.023 \\
DAGMA-L & 0.057 $\pm$ 0.128 & 0.024 $\pm$ 0.053 & 0.033 $\pm$ 0.075 & 22.400 $\pm$ 1.342 & 0.311 $\pm$ 0.019 & 0.933 $\pm$ 0.056 \\
\midrule
GPT-4o & 0.511 $\pm$ 0.026 & 0.518 $\pm$ 0.049 & 0.514 $\pm$ 0.035 & 16.600 $\pm$ 0.894 & 0.231 $\pm$ 0.012 & 0.692 $\pm$ 0.037 \\
\quad + metadata & 0.594 $\pm$ 0.008 & \secondg{0.776 $\pm$ 0.026} & 0.673 $\pm$ 0.015 & 12.800 $\pm$ 0.447 & 0.178 $\pm$ 0.006 & 0.533 $\pm$ 0.019 \\
Qwen3-8B & \best{1.000 $\pm$ 0.000} & 0.059 $\pm$ 0.000 & 0.111 $\pm$ 0.000 & 16.000 $\pm$ 0.000 & 0.222 $\pm$ 0.000 & 0.667 $\pm$ 0.000 \\
\quad + metadata & 0.777 $\pm$ 0.032 & 0.694 $\pm$ 0.026 & \secondg{0.733 $\pm$ 0.027} & \secondg{8.600 $\pm$ 0.894} & \secondg{0.119 $\pm$ 0.012} & \secondg{0.358 $\pm$ 0.037} \\
Qwen3-14B & 0.000 $\pm$ 0.000 & 0.000 $\pm$ 0.000 & 0.000 $\pm$ 0.000 & 17.000 $\pm$ 0.000 & 0.236 $\pm$ 0.000 & 0.708 $\pm$ 0.000 \\
\quad + metadata & 0.200 $\pm$ 0.000 & 0.059 $\pm$ 0.000 & 0.091 $\pm$ 0.000 & 20.000 $\pm$ 0.000 & 0.278 $\pm$ 0.000 & 0.833 $\pm$ 0.000 \\
\midrule
LLM-greedy & 0.015 $\pm$ 0.034 & 0.012 $\pm$ 0.026 & 0.013 $\pm$ 0.030 & 30.400 $\pm$ 1.342 & 0.422 $\pm$ 0.019 & 1.267 $\pm$ 0.056 \\
LLM-CD & 0.715 $\pm$ 0.032 & 0.529 $\pm$ 0.000 & 0.608 $\pm$ 0.011 & 11.600 $\pm$ 0.548 & 0.161 $\pm$ 0.008 & 0.483 $\pm$ 0.023 \\
\midrule
LLM-CR & \secondg{0.787 $\pm$ 0.020} & \best{1.000 $\pm$ 0.000} & \best{0.881 $\pm$ 0.013} & \best{4.600 $\pm$ 0.548} & \best{0.064 $\pm$ 0.008} & \best{0.192 $\pm$ 0.023} \\
\bottomrule
\end{tabular}
}
\caption{Five-run causal discovery performance on the Alzheimer dataset.}
\label{tab:alzheimer-results}
\end{table*}

\begin{table*}[t!]
\centering
\small
\setlength{\tabcolsep}{4pt}

\resizebox{\textwidth}{!}{%
\begin{tabular}{lcccccc}
\toprule
\textbf{Method} & \textbf{Prec.$\uparrow$} & \textbf{Rec.$\uparrow$} & \textbf{F1$\uparrow$} & \textbf{SHD$\downarrow$} & \textbf{NHD$\downarrow$} & \textbf{Ratio$\downarrow$} \\
\midrule
PC & 0.311 $\pm$ 0.050 & 0.175 $\pm$ 0.028 & 0.224 $\pm$ 0.036 & 19.400 $\pm$ 0.894 & 0.346 $\pm$ 0.016 & 1.000 $\pm$ 0.046 \\
GES & \best{0.917 $\pm$ 0.000} & \best{0.688 $\pm$ 0.000} & \best{0.786 $\pm$ 0.000} & \best{6.000 $\pm$ 0.000} & \best{0.107 $\pm$ 0.000} & \best{0.309 $\pm$ 0.000} \\
PC-ANM & 0.381 $\pm$ 0.106 & 0.200 $\pm$ 0.028 & 0.262 $\pm$ 0.048 & 18.200 $\pm$ 1.789 & 0.325 $\pm$ 0.032 & 0.938 $\pm$ 0.092 \\
Direct-LiNGAM & 0.432 $\pm$ 0.075 & \secondg{0.488 $\pm$ 0.068} & 0.458 $\pm$ 0.071 & 18.600 $\pm$ 3.050 & 0.332 $\pm$ 0.054 & 0.959 $\pm$ 0.157 \\
NOTEARS & 0.200 $\pm$ 0.274 & 0.050 $\pm$ 0.068 & 0.080 $\pm$ 0.110 & 18.400 $\pm$ 2.191 & 0.329 $\pm$ 0.039 & 0.948 $\pm$ 0.113 \\
DAG-GNN & 0.067 $\pm$ 0.149 & 0.013 $\pm$ 0.028 & 0.021 $\pm$ 0.047 & 21.200 $\pm$ 2.387 & 0.379 $\pm$ 0.043 & 1.093 $\pm$ 0.123 \\
DAGMA-L & 0.300 $\pm$ 0.274 & 0.075 $\pm$ 0.068 & 0.120 $\pm$ 0.110 & 18.000 $\pm$ 2.739 & 0.321 $\pm$ 0.049 & 0.928 $\pm$ 0.141 \\
\midrule
GPT-4o & 0.212 $\pm$ 0.056 & 0.150 $\pm$ 0.034 & 0.176 $\pm$ 0.043 & 22.600 $\pm$ 1.517 & 0.404 $\pm$ 0.027 & 1.165 $\pm$ 0.078 \\
\quad + metadata & 0.242 $\pm$ 0.028 & 0.275 $\pm$ 0.034 & 0.257 $\pm$ 0.030 & 25.400 $\pm$ 1.140 & 0.454 $\pm$ 0.020 & 1.309 $\pm$ 0.059 \\
Qwen3-8B & 0.714 $\pm$ 0.000 & 0.313 $\pm$ 0.000 & 0.435 $\pm$ 0.000 & 13.000 $\pm$ 0.000 & 0.232 $\pm$ 0.000 & 0.670 $\pm$ 0.000 \\
\quad + metadata & 0.455 $\pm$ 0.000 & 0.313 $\pm$ 0.000 & 0.370 $\pm$ 0.000 & 17.000 $\pm$ 0.000 & 0.304 $\pm$ 0.000 & 0.876 $\pm$ 0.000 \\
Qwen3-14B & 0.500 $\pm$ 0.000 & 0.063 $\pm$ 0.000 & 0.111 $\pm$ 0.000 & 16.000 $\pm$ 0.000 & 0.286 $\pm$ 0.000 & 0.825 $\pm$ 0.000 \\
\quad + metadata & 0.471 $\pm$ 0.039 & 0.225 $\pm$ 0.034 & 0.304 $\pm$ 0.040 & 16.400 $\pm$ 0.548 & 0.293 $\pm$ 0.010 & 0.845 $\pm$ 0.028 \\
\midrule
LLM-greedy & 0.333 $\pm$ 0.000 & 0.188 $\pm$ 0.000 & 0.240 $\pm$ 0.000 & 19.000 $\pm$ 0.000 & 0.339 $\pm$ 0.000 & 0.979 $\pm$ 0.000 \\
LLM-CD &  0.733 $\pm$ 0.091 & 0.275 $\pm$ 0.034 & 0.400 $\pm$ 0.050 & 13.200 $\pm$ 1.095 & 0.236 $\pm$ 0.020 & 0.680 $\pm$ 0.056 \\
\midrule
LLM-CR & 0.679 $\pm$ 0.018 & \best{0.688 $\pm$ 0.000} & \secondg{0.683 $\pm$ 0.009} & \secondg{10.200 $\pm$ 0.447} & \secondg{0.182 $\pm$ 0.008} & \secondg{0.526 $\pm$ 0.023} \\
\bottomrule
\end{tabular}
}
\caption{Five-run causal discovery performance on the Gait dataset.}
\label{tab:gait-results}
\end{table*}

\begin{table*}[t!]
\centering
\small
\setlength{\tabcolsep}{4pt}

\resizebox{\textwidth}{!}{%
\begin{tabular}{lcccccc}
\toprule
\textbf{Method} & \textbf{Prec.$\uparrow$} & \textbf{Rec.$\uparrow$} & \textbf{F1$\uparrow$} & \textbf{SHD$\downarrow$} & \textbf{NHD$\downarrow$} & \textbf{Ratio$\downarrow$} \\
\midrule
PC & 0.600 $\pm$ 0.000 & 0.300 $\pm$ 0.000 & 0.400 $\pm$ 0.000 & 9.000 $\pm$ 0.000 & 0.300 $\pm$ 0.000 & 1.000 $\pm$ 0.000 \\
GES & 0.714 $\pm$ 0.000 & 0.500 $\pm$ 0.000 & 0.588 $\pm$ 0.000 & 7.000 $\pm$ 0.000 & 0.233 $\pm$ 0.000 & 0.778 $\pm$ 0.000 \\
PC-ANM & 0.667 $\pm$ 0.000 & 0.400 $\pm$ 0.000 & 0.500 $\pm$ 0.000 & 8.000 $\pm$ 0.000 & 0.267 $\pm$ 0.000 & 0.889 $\pm$ 0.000 \\
Direct-LiNGAM & 0.700 $\pm$ 0.000 & 0.700 $\pm$ 0.000 & 0.700 $\pm$ 0.000 & 6.000 $\pm$ 0.000 & 0.200 $\pm$ 0.000 & 0.667 $\pm$ 0.000 \\
NOTEARS & 0.250 $\pm$ 0.000 & 0.100 $\pm$ 0.000 & 0.143 $\pm$ 0.000 & 12.000 $\pm$ 0.000 & 0.400 $\pm$ 0.000 & 1.333 $\pm$ 0.000 \\
DAG-GNN & 0.234 $\pm$ 0.023 & 0.140 $\pm$ 0.055 & 0.170 $\pm$ 0.042 & 13.200 $\pm$ 1.304 & 0.440 $\pm$ 0.043 & 1.467 $\pm$ 0.145 \\
DAGMA-L & 0.167 $\pm$ 0.000 & 0.100 $\pm$ 0.000 & 0.125 $\pm$ 0.000 & 14.000 $\pm$ 0.000 & 0.467 $\pm$ 0.000 & 1.556 $\pm$ 0.000 \\
\midrule
GPT-4o & 0.400 $\pm$ 0.548 & 0.040 $\pm$ 0.055 & 0.073 $\pm$ 0.100 & 9.600 $\pm$ 0.548 & 0.320 $\pm$ 0.018 & 1.067 $\pm$ 0.061 \\
\quad + metadata & 0.698 $\pm$ 0.037 & \secondg{0.740 $\pm$ 0.055} & \secondg{0.718 $\pm$ 0.042} & \secondg{5.800 $\pm$ 0.837} & \secondg{0.193 $\pm$ 0.028} & \secondg{0.644 $\pm$ 0.093} \\
Qwen3-8B & 0.750 $\pm$ 0.000 & 0.600 $\pm$ 0.000 & 0.667 $\pm$ 0.000 & 6.000 $\pm$ 0.000 & 0.200 $\pm$ 0.000 & 0.667 $\pm$ 0.000 \\
\quad + metadata & 0.700 $\pm$ 0.000 & 0.700 $\pm$ 0.000 & 0.700 $\pm$ 0.000 & 6.000 $\pm$ 0.000 & 0.200 $\pm$ 0.000 & 0.667 $\pm$ 0.000 \\
Qwen3-14B & 0.000 $\pm$ 0.000 & 0.000 $\pm$ 0.000 & 0.000 $\pm$ 0.000 & 10.000 $\pm$ 0.000 & 0.333 $\pm$ 0.000 & 1.111 $\pm$ 0.000 \\
\quad + metadata & \best{0.783 $\pm$ 0.217} & 0.180 $\pm$ 0.084 & 0.284 $\pm$ 0.110 & 8.800 $\pm$ 0.837 & 0.293 $\pm$ 0.028 & 0.978 $\pm$ 0.093 \\
\midrule
LLM-greedy & 0.414 $\pm$ 0.088 & 0.320 $\pm$ 0.045 & 0.361 $\pm$ 0.061 & 11.400 $\pm$ 1.342 & 0.380 $\pm$ 0.045 & 1.267 $\pm$ 0.149 \\
LLM-CD & 0.500 $\pm$ 0.000 & 0.400 $\pm$ 0.000 & 0.444 $\pm$ 0.000 & 10.000 $\pm$ 0.000 & 0.333 $\pm$ 0.000 & 1.111 $\pm$ 0.000 \\
\midrule
LLM-CR & \secondg{0.773 $\pm$ 0.060} & \best{1.000 $\pm$ 0.000} & \best{0.871 $\pm$ 0.038} & \best{3.000 $\pm$ 1.000} & \best{0.100 $\pm$ 0.033} & \best{0.333 $\pm$ 0.111} \\
\bottomrule
\end{tabular}
}
\caption{Five-run causal discovery performance on the Agro dataset.}
\label{tab:agro-results}
\end{table*}

\begin{table*}[t!]
\centering
\small
\setlength{\tabcolsep}{4pt}

\resizebox{\textwidth}{!}{%
\begin{tabular}{lcccccc}
\toprule
\textbf{Method} & \textbf{Prec.$\uparrow$} & \textbf{Rec.$\uparrow$} & \textbf{F1$\uparrow$} & \textbf{SHD$\downarrow$} & \textbf{NHD$\downarrow$} & \textbf{Ratio$\downarrow$} \\
\midrule
PC & \best{1.000 $\pm$ 0.000} & 0.400 $\pm$ 0.000 & 0.571 $\pm$ 0.000 & 3.000 $\pm$ 0.000 & 0.150 $\pm$ 0.000 & 1.000 $\pm$ 0.000 \\
GES & \best{1.000 $\pm$ 0.000} & 0.400 $\pm$ 0.000 & 0.571 $\pm$ 0.000 & 3.000 $\pm$ 0.000 & 0.150 $\pm$ 0.000 & 1.000 $\pm$ 0.000 \\
PC-ANM & \best{1.000 $\pm$ 0.000} & 0.440 $\pm$ 0.089 & 0.607 $\pm$ 0.080 & 2.800 $\pm$ 0.447 & 0.140 $\pm$ 0.022 & 0.933 $\pm$ 0.149 \\
Direct-LiNGAM & 0.720 $\pm$ 0.073 & \secondg{0.800 $\pm$ 0.000} & 0.756 $\pm$ 0.040 & 2.600 $\pm$ 0.548 & 0.130 $\pm$ 0.027 & 0.867 $\pm$ 0.183 \\
NOTEARS & 0.000 $\pm$ 0.000 & 0.000 $\pm$ 0.000 & 0.000 $\pm$ 0.000 & 5.000 $\pm$ 0.000 & 0.250 $\pm$ 0.000 & 1.667 $\pm$ 0.000 \\
DAG-GNN & 0.000 $\pm$ 0.000 & 0.000 $\pm$ 0.000 & 0.000 $\pm$ 0.000 & 6.800 $\pm$ 0.447 & 0.340 $\pm$ 0.022 & 2.267 $\pm$ 0.149 \\
DAGMA-L & \best{1.000 $\pm$ 0.000} & 0.400 $\pm$ 0.000 & 0.571 $\pm$ 0.000 & 3.000 $\pm$ 0.000 & 0.150 $\pm$ 0.000 & 1.000 $\pm$ 0.000 \\
\midrule
GPT-4o & 0.600 $\pm$ 0.000 & 0.600 $\pm$ 0.000 & 0.600 $\pm$ 0.000 & 4.000 $\pm$ 0.000 & 0.200 $\pm$ 0.000 & 1.333 $\pm$ 0.000 \\
\quad + metadata & 0.571 $\pm$ 0.000 & \secondg{0.800 $\pm$ 0.000} & 0.667 $\pm$ 0.000 & 4.000 $\pm$ 0.000 & 0.200 $\pm$ 0.000 & 1.333 $\pm$ 0.000 \\
Qwen3-8B & 0.667 $\pm$ 0.000 & 0.400 $\pm$ 0.000 & 0.500 $\pm$ 0.000 & 4.000 $\pm$ 0.000 & 0.200 $\pm$ 0.000 & 1.333 $\pm$ 0.000 \\
\quad + metadata & \secondg{0.800 $\pm$ 0.000} & \secondg{0.800 $\pm$ 0.000} & \secondg{0.800 $\pm$ 0.000} & \secondg{2.000 $\pm$ 0.000} & \secondg{0.100 $\pm$ 0.000} & \secondg{0.667 $\pm$ 0.000} \\
Qwen3-14B & \best{1.000 $\pm$ 0.000} & 0.320 $\pm$ 0.110 & 0.476 $\pm$ 0.130 & 3.400 $\pm$ 0.548 & 0.170 $\pm$ 0.027 & 1.133 $\pm$ 0.183 \\
\quad + metadata & \secondg{0.800 $\pm$ 0.000} & \secondg{0.800 $\pm$ 0.000} & \secondg{0.800 $\pm$ 0.000} & \secondg{2.000 $\pm$ 0.000} & \secondg{0.100 $\pm$ 0.000} & \secondg{0.667 $\pm$ 0.000} \\
\midrule
LLM-greedy & 0.480 $\pm$ 0.110 & 0.480 $\pm$ 0.110 & 0.480 $\pm$ 0.110 & 5.200 $\pm$ 1.095 & 0.260 $\pm$ 0.055 & 1.733 $\pm$ 0.365 \\
LLM-CD & 0.333 $\pm$ 0.000 & 0.200 $\pm$ 0.000 & 0.250 $\pm$ 0.000 & 6.000 $\pm$ 0.000 & 0.300 $\pm$ 0.000 & 2.000 $\pm$ 0.000 \\
\midrule
LLM-CR & \best{1.000 $\pm$ 0.000} & \best{0.960 $\pm$ 0.089} & \best{0.978 $\pm$ 0.050} & \best{0.200 $\pm$ 0.447} & \best{0.010 $\pm$ 0.022} & \best{0.067 $\pm$ 0.149} \\
\bottomrule
\end{tabular}
}
\caption{Five-run causal discovery performance on the Coma dataset.}
\label{tab:coma-results}
\end{table*}

\begin{table*}[t!]
\centering
\small
\setlength{\tabcolsep}{4pt}

\resizebox{\textwidth}{!}{%
\begin{tabular}{lcccccc}
\toprule
\textbf{Method} & \textbf{Prec.$\uparrow$} & \textbf{Rec.$\uparrow$} & \textbf{F1$\uparrow$} & \textbf{SHD$\downarrow$} & \textbf{NHD$\downarrow$} & \textbf{Ratio$\downarrow$} \\
\midrule
PC & 0.419 $\pm$ 0.026 & 0.576 $\pm$ 0.026 & 0.485 $\pm$ 0.026 & 20.800 $\pm$ 1.304 & 0.189 $\pm$ 0.012 & 1.000 $\pm$ 0.063 \\
GES & 0.154 $\pm$ 0.036 & 0.282 $\pm$ 0.064 & 0.199 $\pm$ 0.046 & 38.600 $\pm$ 2.608 & 0.351 $\pm$ 0.024 & 1.856 $\pm$ 0.125 \\
PC-ANM & 0.419 $\pm$ 0.026 & 0.576 $\pm$ 0.026 & 0.485 $\pm$ 0.026 & 20.800 $\pm$ 1.304 & 0.189 $\pm$ 0.012 & 1.000 $\pm$ 0.063 \\
Direct-LiNGAM & 0.285 $\pm$ 0.015 & 0.576 $\pm$ 0.026 & 0.381 $\pm$ 0.018 & 31.800 $\pm$ 1.304 & 0.289 $\pm$ 0.012 & 1.529 $\pm$ 0.063 \\
NOTEARS & 0.373 $\pm$ 0.037 & 0.118 $\pm$ 0.000 & 0.179 $\pm$ 0.004 & 18.400 $\pm$ 0.548 & 0.167 $\pm$ 0.005 & 0.885 $\pm$ 0.026 \\
DAG-GNN & 0.293 $\pm$ 0.055 & 0.165 $\pm$ 0.026 & 0.211 $\pm$ 0.035 & 21.000 $\pm$ 1.225 & 0.191 $\pm$ 0.011 & 1.010 $\pm$ 0.059 \\
DAGMA-L & 0.270 $\pm$ 0.048 & 0.129 $\pm$ 0.026 & 0.174 $\pm$ 0.032 & 20.800 $\pm$ 1.095 & 0.189 $\pm$ 0.010 & 1.000 $\pm$ 0.053 \\
\midrule
GPT-4o & 0.252 $\pm$ 0.014 & 0.447 $\pm$ 0.032 & 0.322 $\pm$ 0.019 & 32.000 $\pm$ 0.707 & 0.291 $\pm$ 0.006 & 1.538 $\pm$ 0.034 \\
\quad + metadata & 0.226 $\pm$ 0.007 & 0.353 $\pm$ 0.000 & 0.275 $\pm$ 0.006 & 31.600 $\pm$ 0.894 & 0.287 $\pm$ 0.008 & 1.519 $\pm$ 0.043 \\
Qwen3-8B & 0.500 $\pm$ 0.000 & 0.118 $\pm$ 0.000 & 0.190 $\pm$ 0.000 & 17.000 $\pm$ 0.000 & 0.155 $\pm$ 0.000 & 0.817 $\pm$ 0.000 \\
\quad + metadata & 0.533 $\pm$ 0.000 & 0.471 $\pm$ 0.000 & \secondg{0.500 $\pm$ 0.000} & \secondg{16.000 $\pm$ 0.000} & \secondg{0.145 $\pm$ 0.000} & \secondg{0.769 $\pm$ 0.000} \\
Qwen3-14B & \secondg{0.562 $\pm$ 0.069} & 0.235 $\pm$ 0.000 & 0.331 $\pm$ 0.012 & 16.200 $\pm$ 0.837 & 0.147 $\pm$ 0.008 & 0.779 $\pm$ 0.040 \\
\quad + metadata & 0.478 $\pm$ 0.030 & 0.235 $\pm$ 0.000 & 0.315 $\pm$ 0.007 & 17.400 $\pm$ 0.548 & 0.158 $\pm$ 0.005 & 0.837 $\pm$ 0.026 \\
\midrule
LLM-greedy & 0.400 $\pm$ 0.000 & \secondg{0.588 $\pm$ 0.000} & 0.476 $\pm$ 0.000 & 22.000 $\pm$ 0.000 & 0.200 $\pm$ 0.000 & 1.058 $\pm$ 0.000 \\
LLM-CD & 0.405 $\pm$ 0.018 & 0.576 $\pm$ 0.026 & 0.476 $\pm$ 0.021 & 21.600 $\pm$ 0.894 & 0.196 $\pm$ 0.008 & 1.038 $\pm$ 0.043 \\
\midrule
LLM-CR & \best{0.600 $\pm$ 0.000} & \best{0.882 $\pm$ 0.000} & \best{0.714 $\pm$ 0.000} & \best{12.000 $\pm$ 0.000} & \best{0.109 $\pm$ 0.000} & \best{0.577 $\pm$ 0.000} \\
\bottomrule
\end{tabular}
}
\caption{Five-run causal discovery performance on the Sachs dataset.}
\label{tab:sachs-results}
\end{table*}

\begin{table}[t]
\centering
\small
\setlength{\tabcolsep}{3.5pt}

\begin{tabular}{@{}lccccc@{}}
\toprule
Graph & CP & Gait & Agro & Alz. & Coma \\
\midrule
Data-driven graph       & 0.502 & \textbf{0.786} & 0.588 & 0.335 & 0.571 \\
Verified LLM graph      & 0.783 & 0.457 & 0.870 & 0.750 & 0.769 \\
Shared edges only       & 0.692 & 0.545 & 0.333 & 0.538 & 0.571 \\
After CR                & \textbf{0.805} & 0.710 & 0.667 & 0.733 & 0.889 \\
Full LLM-CR             & \textbf{0.805} & 0.683 & \textbf{0.871} &
                          \textbf{0.881} & \textbf{0.978} \\
\midrule

Graph union             & 0.651 & 0.634 & 0.800 & 0.791 & 0.769 \\
Rescore all union edges & 0.621 & 0.786 & 0.400 & 0.643 & 0.750 \\

\bottomrule
\end{tabular}
\caption{Stage-wise and graph-combination ablations.}
\label{tab:stage_ablation}
\end{table}

Tables~\ref{tab:cp-results}--\ref{tab:sachs-results} present detailed comparisons with the strongest competing baseline on each dataset. LLM-CR improves recall on the majority of datasets, while matching the recall of GES on Gait. This increased sensitivity involves a precision trade-off. Overall, compared with the strongest dataset-specific baselines, LLM-CR increases macro-average recall by 27.0\%, reduces average SHD by 40.5\%, and improves macro-average F1 by 17.0\%. These results indicate that LLM-CR prioritizes recovering true causal relations while maintaining a favorable overall balance between edge-recovery performance and structural error.

\label{detailed results}
\subsection{Stage-wise Ablation}

Table~\ref{tab:stage_ablation} reports F1 for the data-driven graph, verified LLM graph, their shared edges, and two stages of LLM-CR: conflict resolution before recovery (“After CR”) and the full pipeline after recovery. Two additional controls: the union of both graphs and rescoring all edges in their union, including shared edges. Conflict resolution improves on the verified LLM graph for CP, Gait, and Coma; recovery produces the main gains for Agro, Coma, and Alzheimer, leaves CP unchanged, and lowers F1 for Gait. The full pipeline exceeds both the verified LLM graph and graph union on all five benchmarks, although GES remains stronger on Gait.
\subsection{Alternative Data-driven Backbones}
\begin{table*}[t]
\centering
\small

\setlength{\tabcolsep}{4pt}
\resizebox{\textwidth}{!}{%
\begin{tabular}{ll*{10}{c}}
\toprule
\multirow{2}{*}{\textbf{Backbone}} & \multirow{2}{*}{\textbf{Graph}}
& \multicolumn{2}{c}{\textbf{CP}}
& \multicolumn{2}{c}{\textbf{Gait}}
& \multicolumn{2}{c}{\textbf{Agro}}
& \multicolumn{2}{c}{\textbf{Alzheimer}}
& \multicolumn{2}{c}{\textbf{Coma}} \\
\cmidrule(lr){3-4}\cmidrule(lr){5-6}
\cmidrule(lr){7-8}\cmidrule(lr){9-10}
\cmidrule(lr){11-12}
& & F1$\uparrow$ & NHD$\downarrow$
& F1$\uparrow$ & NHD$\downarrow$
& F1$\uparrow$ & NHD$\downarrow$
& F1$\uparrow$ & NHD$\downarrow$
& F1$\uparrow$ & NHD$\downarrow$ \\
\midrule
GES & Data only & 0.502 & 0.282 & 0.786 & 0.107 & 0.588 & 0.233 & 0.335 & 0.300 & 0.571 & 0.150 \\
 & LLM-CR & 0.805 & 0.115 & 0.683 & 0.182 & 0.871 & 0.100 & 0.881 & 0.064 & 0.978 & 0.010 \\
\midrule
PC & Data only & 0.183 & 0.318 & 0.224 & 0.346 & 0.400 & 0.300 & 0.089 & 0.333 & 0.571 & 0.150 \\
 & LLM-CR & 0.667 & 0.155 & 0.533 & 0.250 & 0.842 & 0.100 & 0.743 & 0.125 & 0.888 & 0.050\\
\midrule
DAG-GNN & Data only & 0.000 & 0.273 & 0.021 & 0.379 & 0.170 & 0.440 & 0.109 & 0.314 & 0.000 & 0.340 \\
 & LLM-CR & 0.740 & 0.127 & 0.518 & 0.232 & 0.667 & 0.200 & 0.709 & 0.125 & 0.571 & 0.150 \\
\bottomrule
\end{tabular}%
}
\caption{Effect of the data-driven backbone on graph recovery across five datasets.}
\label{tab:backbone_ablation}
\end{table*} Table~\ref{tab:backbone_ablation} reports each backbone’s graph alone and the final LLM-CR graph obtained from it. This comparison shows how backbone quality affects the reconciled graph and tests whether the integration procedure transfers beyond GES.
\label{appendix:H}

\section{Additional Methodological Details}
\label{sec: appendix c}
\subsection{BIC-based Data Fit Component}
For a candidate edge \(e = X_i \rightarrow X_j\), we start from the common edge graph \(G_c\), which contains edges ($\mathcal{E}_{\mathrm{common}}$) supported by both data-driven and verified LLM-driven causal discovery. Then we compute the local BIC gain by comparing the score of the target variable \(X_j\) before and after adding \(X_i\) to its parent set. We define the BIC score gain as:
\begin{equation}
\begin{aligned}
\Delta \mathrm{BIC}(e \mid G_c)
&=
\mathrm{BIC}\!\left(
X_j \mid \mathrm{Pa}_{G_c}(X_j) \cup \{X_i\}
\right) \\
&\quad -
\mathrm{BIC}\!\left(
X_j \mid \mathrm{Pa}_{G_c}(X_j)
\right)
\end{aligned}
\end{equation}
Here, \(G_c\) is the current graph, and \(\mathrm{Pa}_{G_c}(X_j)\) denotes the existing parent set of \(X_j\). 
In our implementation, we use the higher-is-better BIC convention, where the score is defined as the log-likelihood minus a model-complexity penalty. A positive \(\Delta \mathrm{BIC}(e \mid G_c)\) indicates that the candidate edge improves the graph after adding the parent, whereas a negative value indicates that the edge worsens the model fit or introduces unnecessary complexity. This allows BIC to act as a data-driven structural fitness term during disputed-edge resolution.
\subsection{HSIC-based Dependence Component}
HSIC measures whether adding the candidate parent reduces the non-linear dependence between the candidate parent and the unexplained residual of the child variable. Starting from the common edge graph \(G_c\), we consider each disputed candidate edge \(e = X_i \rightarrow X_j\) and fit a local linear model for \(X_j\) using its existing parents and compute the residuals. Next, we add the candidate parent \(X_i\) to the parent set, refit the local model, and compute the new residuals. HSIC is then used to measure the nonlinear dependence between \(X_i\) and the residuals before and after adding the candidate edge. Let \(G_c\) denote the common-edge graph and let \(e = X_i \rightarrow X_j\) be a disputed candidate edge.
The existing parent set of \(X_j\) is denoted as \(\mathrm{Pa}_{G_c}(X_j)\).
We first compute the residual \(r_j\) of \(X_j\) using its current parents:

\begin{equation}
r_j^{-e}
=
X_j
-
\hat{f}\left(\mathrm{Pa}_{G_c}(X_j)\right)
\end{equation}

Here, \(r_j^{-e}\) represents the residual state without edge \(e\). After adding the candidate parent \(X_i\), the updated residual is:

\begin{equation}
r_j^{+e}
=
X_j
-
\hat{f}\left(\mathrm{Pa}_{G_c}(X_j) \cup \{X_i\}\right)
\end{equation}

Here, \(r_j^{+e}\) represents the residual state with edge \(e\). The HSIC gain is then defined as:

\begin{equation}
\begin{aligned}
\Delta \mathrm{HSIC}(e \mid G_c)
&= \mathrm{HSIC}\!\left(X_i, r_j^{-e}\right) \\
&\quad - \mathrm{HSIC}\!\left(X_i, r_j^{+e}\right)
\end{aligned}
\end{equation}
If adding \(X_i\) as a parent substantially reduces this dependence, the candidate edge receives a positive HSIC gain. Therefore, the HSIC component acts as a nonlinear residual dependence-reduction term, supporting edges that explain the remaining dependence not captured by the current parent set. In our implementation, HSIC is efficiently approximated using Random Fourier Features~\cite{zhang2018large} with an RBF kernel, and the HSIC component retains candidate edges only when it yields an improvement. 

\subsection{Semantic Entropy Calculation}
\label{sec:ap_se}
Semantic entropy is used in our framework to quantify the uncertainty of LLM-based causal judgments beyond simple majority voting. For each valid response \(r\) from the response generator, the judgment and its one-sentence causal explanation are combined into a semantic representation for the directed pair. Responses are then grouped into semantic clusters using bidirectional entailment: two responses are assigned to the same cluster only when the DeBERTa-large-MNLI~\cite{he2021deberta} model predicts that each response entails the other. Responses that lack a judgment, an explanation sentence, or the required explanation log-probability are excluded from the entropy computation. This procedure ensures that differently worded responses expressing the same causal interpretation are treated as one semantic outcome, while responses expressing substantively different explanations remain in separate clusters.
Following~\cite{farquhar2024detecting}, semantic entropy is computed over meanings rather than surface-level generations. For each candidate edge $e$, we sample $R$ responses $\mathcal{S}(e)=\{s^{(1)},\ldots,s^{(R)}\}$ from the LLM.
For a generated response $s^{(r)}=(s^{(r)}_1,\ldots,s^{(r)}_{T_r})$, its sequence log-probability is computed as
\begin{equation}
\log \tilde{P}(s^{(r)} \mid e)
=
\frac{1}{T_r}
\sum_{t=1}^{T_r}
\log P\!\left(s^{(r)}_t \mid s^{(r)}_{<t}, e\right)
\end{equation}
Following semantic entropy, we group the sampled responses into semantic
equivalence classes using bidirectional entailment. Let
$\mathcal{C}(e)=\{C_1,\ldots,C_K\}$ denote the resulting set of semantic
clusters for edge $e$. The probability mass of a semantic cluster $C_k$ is
estimated by summing the probabilities of all responses assigned to that
cluster:
\begin{equation}
P(C_k \mid e)
=
\frac{
\sum_{s^{(r)} \in C_k} P(s^{(r)} \mid e)
}{
\sum_{j=1}^{K} \sum_{s^{(r)} \in C_j} P(s^{(r)} \mid e)
}
\end{equation}
The semantic entropy of edge $e$ is then computed over the normalized
distribution of semantic clusters:
\begin{equation}
\mathrm{SE}(e)
=
-\sum_{k=1}^{K}
P(C_k \mid e)\log P(C_k \mid e)
\end{equation}

We use normalized semantic entropy to calculate the weight, or confidence, of the LLM's response.
\begin{equation}
\mathrm{SE}_{\mathrm{norm}}(e)
=
\frac{\mathrm{SE}(e)}
{\log |\mathcal{C}(e)| + \epsilon}
\end{equation}
When all
responses are semantically consistent, $\mathrm{SE}_{\mathrm{norm}}(e)$ is low;
when the responses are semantically diverse or contradictory,
$\mathrm{SE}_{\mathrm{norm}}(e)$ is high.
\subsection{Uncertainty-aware LLM Prior Component}
The BIC and HSIC components evaluate whether a candidate edge is supported by the observed data, but statistical evidence alone may be insufficient when two directions are statistically difficult to distinguish, or the sample size is limited. We therefore introduce an uncertainty-aware LLM prior to provide semantic evidence about whether a disputed directed edge \(e=X_i\rightarrow X_j\) is causally plausible. This component is not intended to replace the data-driven criteria. Instead, it complements them by contributing knowledge derived from variable meanings, metadata, and the causal explanations generated by the LLM. 
For each directed edge, we collect multiple verified LLM responses by the response generator component in LLM-driven graph construction, and convert each valid response into a binary plausibility value, assigning 1 to plausible responses and 0 to implausible and uncertain ones (Figure~\ref{fig:framework}(b)). As we have generated $R$ responses for each pair, for each candidate directed edge $e_{ij}: X_i \rightarrow X_j$, the raw LLM plausibility is computed as the proportion of valid responses that support the edge.
\begin{equation}
p_{ij}
=
\frac{N_{\mathrm{plausible}}(e)}
{N_{\mathrm{responses}}(e)} 
\end{equation}
where \(N_{\mathrm{plausible}}(e)\) is the number of valid responses supporting (e), and \(N_{\mathrm{responses}}(e)\) is the total number of valid plausible, implausible and uncertain responses. For example, if eight of ten valid responses classify \(X_i\rightarrow X_j\) as plausible, then \(p_{\mathrm{LLM}}(e)=0.8\). Although this proportion captures agreement among repeated generations, it does not indicate whether the responses express one consistent causal explanation or several conflicting interpretations. To measure this additional uncertainty, we calculate semantic entropy at the level of response meanings rather than individual output tokens~\cite{farquhar2024detecting}.
We define the LLM reliability weight as
\begin{equation}
w_{ij}
=
1 - SE_{ij}^{\mathrm{norm}} 
\end{equation}

When semantic entropy is high, $w_{ij}$ is small and $p_{ij}^{\mathrm{adj}}$ is shrunk toward a sparse baseline $\pi$, thereby preventing highly uncertain LLM judgments from dominating the conflict resolution score. The uncertainty-adjusted LLM probability is
\begin{equation}
p_{ij}^{\mathrm{adj}}
= (1-w_{ij})\pi + w_{ij}p_{ij}^{\mathrm{LLM}},
\end{equation}
or,
equivalently,
\begin{equation}
p_{ij}^{\mathrm{adj}}
= \pi + w_{ij}\left(p_{ij}^{\mathrm{LLM}}-\pi\right).
\end{equation}
This formulation satisfies two boundary conditions:
\begin{equation}
\begin{aligned}
w_{ij} = 0 &\implies p_{ij}^{\mathrm{adj}} = \pi, \\
w_{ij} = 1 &\implies p_{ij}^{\mathrm{adj}} = p_{ij}^{\mathrm{LLM}}.
\end{aligned}
\end{equation}
Therefore, uncertain LLM judgments are pulled toward the baseline prior,
while confident LLM judgments remain close to the LLM plausibility score.

Before converting the adjusted probability into a log-odds score, we clip
it for numerical stability:
\begin{equation}
\tilde{p}_{ij}^{\mathrm{adj}}
=
\mathrm{clip}
\left(
p_{ij}^{\mathrm{adj}},
\epsilon,
1-\epsilon
\right)
\end{equation}

The LLM prior score for edge $X_i \rightarrow X_j$ is then defined as the
log-odds of the adjusted probability:
\begin{equation}
\mathrm{Prior}_{\mathrm{LLM}}(e_{ij})
=
\log
\frac{
\tilde{p}_{ij}^{\mathrm{adj}}
}{
1-\tilde{p}_{ij}^{\mathrm{adj}}
}
\end{equation}

A positive value indicates that the uncertainty-aware LLM evidence
supports the edge more than its absence, while a negative value indicates
weak LLM support.
\subsection{Post-hoc Sensitivity Analysis of Hyperparameters}
~\label{senstivity_analysis}
\begin{figure}[t]
    \centering
    \includegraphics[width=\columnwidth]{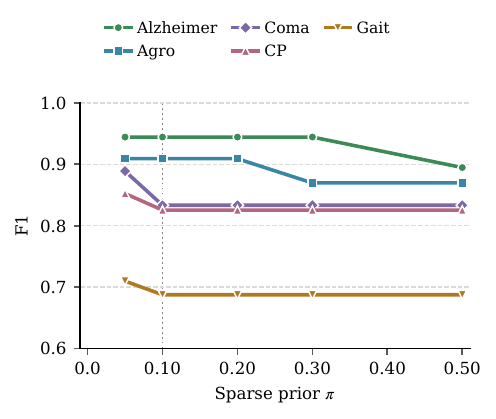}
    \caption{Sensitivity of F1 to the sparse prior $\pi$ across five benchmarks.}
    \label{fig:sparse-prior-sensitivity}
\end{figure}

\begin{figure}[t]
    \centering
    \includegraphics[width=\columnwidth]{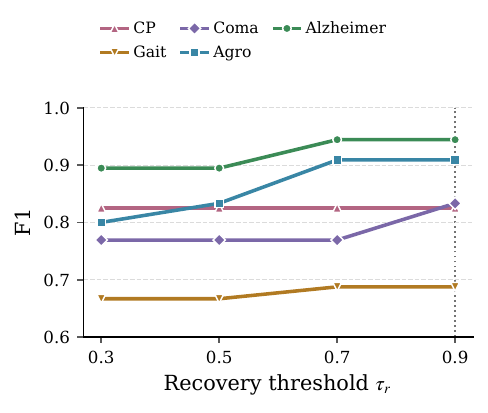}
    \caption{Sensitivity of F1 scores to the recovery
    threshold $\tau_r$ across datasets.}
    \label{fig:recovery_threshold}
\end{figure}

\begin{table}[t]
\centering
\footnotesize
\setlength{\tabcolsep}{6pt}

\begin{tabular}{lccc}
\toprule
Dataset & \shortstack{Fixed\\F1}
        & \shortstack{Post-hoc\\max. F1}
        & \shortstack{Weights at max. F1} \\
\midrule
CP        & 0.805 & 0.839 & $(0.35,0.20,0.45)$ \\
Gait      & 0.683 & 0.759 & $(1.00,0.00,0.00)$ \\
Agro      & 0.871 & 0.947 & $(0.60,0.30,0.10)$ \\
Alzheimer & 0.881 & 0.881 & $(0.40,0.25,0.35)$ \\
Coma      & 0.978 & 0.978 & $(0.40,0.25,0.35)$ \\
\bottomrule
\end{tabular}
\caption{Sensitivity to conflict-resolution weights.}
\label{tab:weight_sensitivity_all}
\end{table}
Figure~\ref{fig:sparse-prior-sensitivity} and~\ref{fig:recovery_threshold}
present a post-hoc sensitivity analysis of the sparse prior
and recovery threshold using a single run of LLM-CR. Table ~\ref{tab:weight_sensitivity_all} compares the best weight settings with the fixed weights used in LLM-CR. We therefore do not select weights by maximizing performance separately on each benchmark, as doing so would tune the framework directly to the evaluation ground truth and could reduce its generalizability. Instead, we selected all hyperparameters using the Sachs dataset and apply the same fixed settings across datasets. Thus, the selected configuration represents a stable and methodologically motivated compromise rather than a dataset-specific optimum.

\begin{figure*}[t]
  \includegraphics[width=0.48\linewidth]{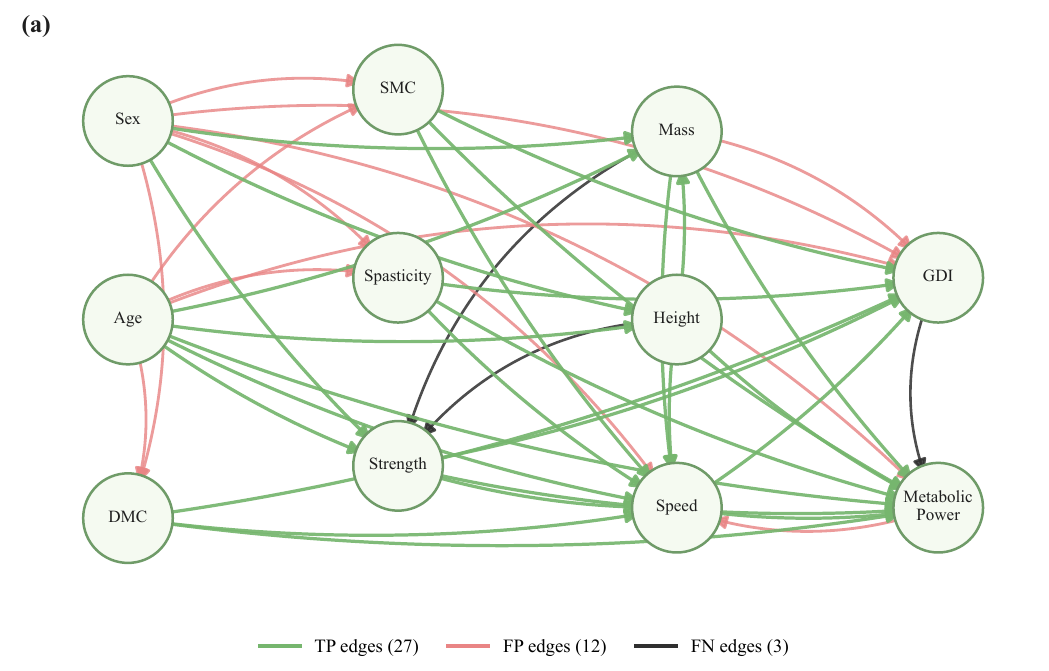} \hfill
  \includegraphics[width=0.48\linewidth]{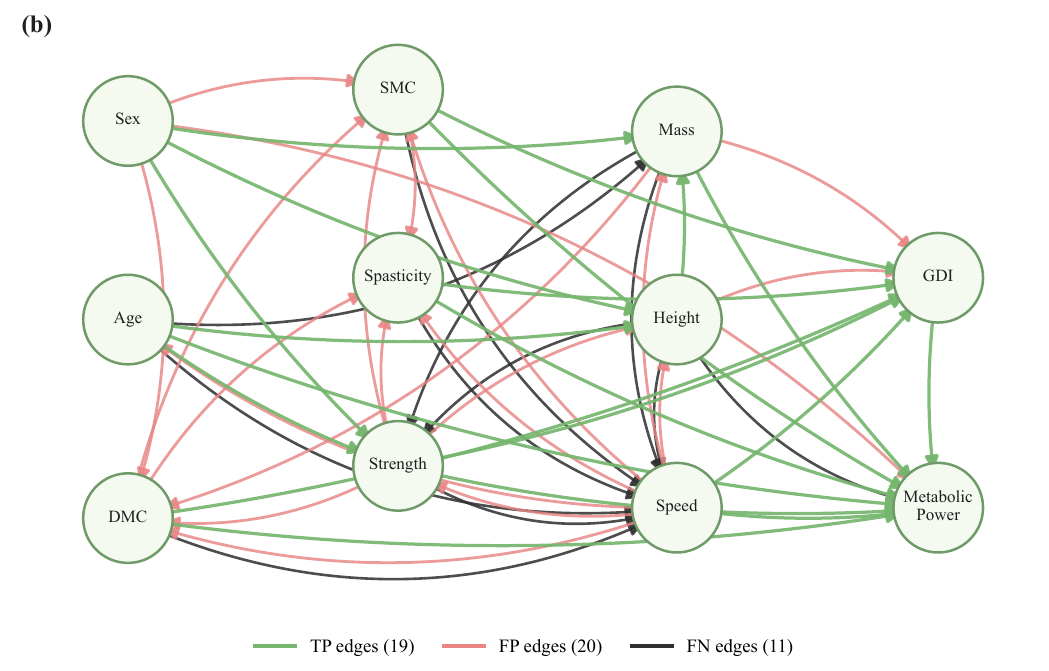}
  \caption{Comparison of the LLM-driven graph (a) and data-driven graph (b) for the CP dataset.}
  \label{fig::Comparison}
\end{figure*}
\begin{figure}[t]
  \includegraphics[width=\columnwidth]{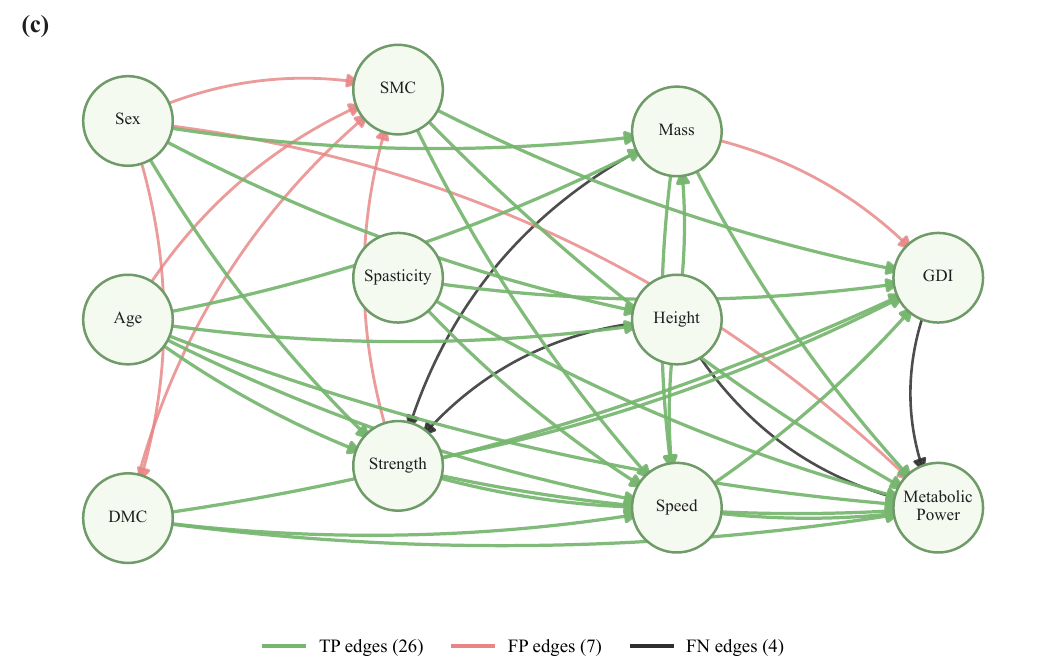}
  \caption{Final graph after edge resolution for the CP dataset.}
  \label{fig:final graph}
\end{figure}

\section{Details of Experimental Setting}

\subsection{Implementation Details}
We use GES~\cite{chickering2002optimal} as the data-driven backbone and GPT-4o as the default LLM for causal discovery, unless otherwise stated.
In the response generator, we generated 10 responses per variable pair with temperature set to 0.1.
The schema generator and verifier use deterministic decoding with temperature 0.0 and produce one response per pair. Moreover, the majority threshold \(\tau_m\) is set to 7 out of 10.
The coefficients in the uncertainty-aware conflict resolution score were 0.4 for BIC, 0.35 for the LLM prior, and 0.25 for the HSIC component, whereas the sparse baseline prior \(\pi\) in the LLM prior was 0.1. BIC receives the largest weight because it directly evaluates whether adding a parent improves local data fit while penalizing unnecessary model complexity. The LLM prior receives the second-largest weight because it provides pair-specific semantic evidence, but its contribution is reduced when semantic entropy is high. HSIC is used as complementary evidence of nonlinear dependence and therefore receives a smaller weight. For LLM-based recovery,  \(\tau_r\) is set to 0.9. 

For the MIMIC, we performed downstream prediction following~\cite{du2025causal}. Moreover, we used stratified five-fold cross-validation. For each fold, the entire graph-construction pipeline was fitted using only the outer training partition, and the final performance was evaluated for the held-out partition. The reported values are the mean and standard deviation across the five held-out folds.
\subsection{Datasets}
We evaluate the proposed framework against baseline methods on seven datasets spanning synthetic benchmarks, real-world biomedical data, and clinical prediction. Coma~\citep{cooper1984nestor} is a widely used Bayesian network, Agro~\citep{baudrit2022decision} models decision support for skimmed-milk microfiltration, and Gait~\citep{steele2022causal} represents a cerebral-palsy-related causal benchmark. For these datasets, observational samples are synthetically generated from the available ground-truth DAGs following established simulation procedures~\citep{yu2019dag,nagesh2026evaluating}. We additionally use the synthetic Alzheimer’s dataset introduced by~\citet{nagesh2026evaluating}, which was generated from a known causal graph. Moreover, we use the Sachs dataset~\cite{sachs2005causal} from \texttt{bnlearn}~\cite{scutari2010learning}. Because Sachs~\cite{sachs2005causal} is widely available in the literature and may be represented in LLM training data, we treat it as potentially contaminated. It was used only during preliminary method development and is excluded from the final benchmark evaluation and all principal conclusions. The final framework configuration was fixed before evaluation on the remaining datasets.
We also use a real de-identified biomedical dataset containing information about the gait of patients with Cerebral Palsy~\cite{gill2023causal}, whose ground-truth DAG was specified by clinical experts, and a lung cancer dataset derived from MIMIC-IV~\cite{johnson2023mimic}. Since the lung cancer dataset does not include a ground-truth DAG, we evaluate the learned graph through downstream prediction using the parents of the target node following~\cite{du2025causal}. The MIMIC-IV cohort comprises 2,330 positive (50.32\%) and 2,300 negative (49.68\%) cases, resulting in an approximately balanced outcome distribution. \\
\label{appen_datsets}

\subsection{Model Details}\label{appendix:G}

\begin{itemize}
  \item \textbf{PC~\cite{spirtes2001causation}:} PC algorithm is a classical constraint-based method that learns graph structure from conditional independencies.
  \item \textbf{GES~\cite{chickering2002optimal}:} Greedy equivalence search (GES) is a score-based algorithm that searches over equivalence classes of DAGs and chooses the DAG with the best score. It greedily adds and removes edges to optimize the BIC score.
  \item \textbf{PC-ANM~\cite{peters2014causal}:} PC-ANM is an extension of functional causal models that first applies PC to estimate the skeleton and then uses an additive noise model to orient the remaining undirected edges.
  \item \textbf{Direct-LiNGAM~\cite{shimizu2011directlingam}:} It is a functional causal method that identifies causal ordering and edge strengths by assuming that variables are generated by linear functions of their causes with independent non-Gaussian noise.
  \item \textbf{NOTEARS~\cite{zheng2018dags}:} NOTEARS formulates causal graph learning as a continuous optimization problem over a weighted adjacency matrix and enforces acyclicity through a smooth constraint.
  \item \textbf{DAG-GNN~\cite{yu2019dag}:} It uses graph neural networks to learn non-linear causal relationships among variables and jointly estimates the graph structure and data-generation process while enforcing acyclicity.
  \item \textbf{DAGMA-L~\cite{bello2022dagma}: } It is a linear version of DAGMA and uses a log-determinant acyclicity constraint to improve optimization behavior.
  \item \textbf{FCI~\cite{spirtes2013causal}:} A constraint-based algorithm designed for settings with potential latent confounders and selection bias.
  \item \textbf{Qwen3}~\cite{yang2025qwen3}: An open-weight LLM family used as LLM-only causal discovery baselines. We evaluate Qwen3-8B and Qwen3-14B under the same prompting setting.
  \item \textbf{LLM-greedy~\cite{long2023causal}:} LLM-greedy treats the language model as an imperfect expert for causal discovery. It greedily incorporates knowledge from the LLM while enforcing constraints such as acyclicity.
  \item \textbf{LLM-CD~\cite{du2025causal}}: LLM-CD is a hybrid causal discovery framework that integrates LLM-based semantic reasoning with data-driven causal discovery. It uses LLM knowledge to support graph construction, edge refinement, and causal relation inference.
  \item \textbf{LLM-BFS~\cite{jiralerspong2024efficient}:} It uses a breadth-first search strategy with LLM queries to construct graphs more efficiently by reducing the number of causal queries and relying on LLM-based causal judgments.
  \item \textbf{ChatPC~\cite{cohrs2023large}:} ChatPC uses an LLM as a conditional-independence oracle within a PC-style causal discovery procedure. It prompts the LLM to answer conditional independence queries and then uses those answers to remove and orient edges.
\end{itemize}

\subsection{Metrics}
\textbf{SHD:} Structural Hamming Distance measures the edge additions and deletions required to transform the learned graph into the ground-truth graph. Lower SHD indicates a graph closer to the true graph.\\
\textbf{NHD:} Normalized Hamming Distance is the normalized version of the SHD, usually obtained by dividing the number of possible or true graph differences by the total number of possible graph differences. We used the number of possible edges to calculate NHD. Lower NHD indicates better structural recovery while accounting for graph size.\\
\textbf{Ratio:} We calculate the ratio to assess our framework's behavior relative to a baseline. For each method, the reported Ratio mean is the method's mean NHD divided by PC's mean NHD. The reported Ratio standard deviation is the method's NHD standard deviation divided by PC's mean NHD. The ratio metrics indicate that lower values are better for higher structural accuracy.\\
\label{metrics details}

\subsection{Runtime and Token Cost}
\label{app:efficiency}
\begin{table}[t]
\centering
\small

\setlength{\tabcolsep}{5pt} 
\begin{tabular}{@{}lrrrrrr@{}}
\toprule
Dataset & Calls & Tokens & LLM & CR & Total & Cost \\
\midrule
CP        & 1211 & 4.26 & 6.39 & 3.29 & 9.68 & 10.90 \\
Gait      &  617 & 1.84 & 4.69 & 1.08 & 5.77 &  4.82 \\
Agro      &  331 & 0.84 & 2.48 & 0.51 & 2.99 &  2.26 \\
Alzheimer &  793 & 2.79 & 6.18 & 1.18 & 7.36 &  7.17 \\
Coma      &  221 & 0.55 & 1.80 & 1.11 & 2.91 &  1.46 \\
MIMIC-IV  & 3366 & 6.94 & 5.42 & 3.95 & 9.37 & 17.80 \\
\bottomrule
\end{tabular}
\caption{Average resource use per run over five runs. \textbf{LLM} is the wall time for constructing the verified LLM graph; \textbf{CR} is the wall time for conflict resolution and recovery; and \textbf{Total} is the sum of the two. Times are in minutes, tokens are in millions, and cost is estimated in USD per run.}
\label{tab:efficiency}
\end{table}
Table~\ref{tab:efficiency} reports the average over five runs. LLM graph was constructed by accessing GPT-4o through the API. Because the LLM branch evaluates ordered variable
pairs, its call volume grows quadratically with the number of variables.

\begin{algorithm}[t]
\small
\caption{Verified LLM-Driven Graph}
\label{alg:llm-driven-graph}
\begin{algorithmic}[1]
\Require Dataset \(\mathcal{D}\), metadata $M$, LLM $f_{\mathrm{LLM}}$, responses per pair $R$, threshold $\tau_m$
\Ensure Verified LLM-driven graph $G_{\mathrm{LLM}}$

\State Generate schema $C$ from \(\mathcal{D}\) and $M$ using $f_{\mathrm{LLM}}$.

\State Construct ordered pairs $P = \{(X_i, X_j): X_i, X_j \in \mathcal{V},\; X_i \neq X_j\}$.

\ForAll{$(X_i, X_j) \in P$}
    \State Generate $R$ judgments for $X_i \rightarrow X_j$.
    \State Map judgments: \texttt{plausible}, \texttt{not\_plausible}, or \texttt{uncertain}.
    \State Assign preliminary label by majority threshold $\tau_m$.
\EndFor

\ForAll{$(X_i, X_j) \in P$}
    \State Verify the preliminary label using $C$ and $M$
    \State Obtain $\mathrm{corrected\_outcome}(X_i \rightarrow X_j)$.
\EndFor

\State $\mathcal{E}_{\mathrm{LLM}} \gets \{X_i \rightarrow X_j : \mathrm{corrected\_outcome}(X_i \rightarrow X_j) = \texttt{plausible}\}$
\State \Return $G_{\mathrm{LLM}} = (\mathcal{V}, \mathcal{E}_{\mathrm{LLM}})$
\end{algorithmic}
\end{algorithm}
\label{Algo:LLM_driven_part}
\begin{algorithm}[t]
\small
\caption{Uncertainty-Aware Conflict Resolution}
\label{alg:conflict-resolution}
\begin{algorithmic}[1]

\Require LLM-driven graph
$G_{\mathrm{LLM}}=(\mathcal{V},\mathcal{E}_{\mathrm{LLM}})$,
data-driven graph
$G_{\mathrm{Data}}=(\mathcal{V},\mathcal{E}_{\mathrm{Data}})$,
dataset \(\mathcal{D}\), LLM prior $\mathrm{Prior}_{\mathrm{LLM}}(e)$,
weights $\lambda_{\mathrm{BIC}},\beta_{\mathrm{HSIC}},\gamma_{\mathrm{LLM}}$,

\Ensure Resolved acyclic graph
$G=(\mathcal{V},\mathcal{E})$

\State $\mathcal{E}_{\mathrm{common}}
\gets \mathcal{E}_{\mathrm{LLM}} \cap \mathcal{E}_{\mathrm{Data}}$

\State $\mathcal{E}_{\mathrm{LLM\text{-}only}}
\gets \mathcal{E}_{\mathrm{LLM}} \setminus \mathcal{E}_{\mathrm{Data}}$

\State $\mathcal{E}_{\mathrm{Data\text{-}only}}
\gets \mathcal{E}_{\mathrm{Data}} \setminus \mathcal{E}_{\mathrm{LLM}}$

\State $\mathcal{E}_{\mathrm{disputed}}
\gets \mathcal{E}_{\mathrm{LLM\text{-}only}}
\cup \mathcal{E}_{\mathrm{Data\text{-}only}}$

\State $\mathcal{E} \gets \mathcal{E}_{\mathrm{common}}$

\ForAll{$e=(X_i \rightarrow X_j)\in \mathcal{E}_{\mathrm{disputed}}$}

    \State $\mathcal{P}_j
    \gets \mathrm{Pa}_{\mathcal{E}_{\mathrm{common}}}(X_j)$

    \State Compute local BIC gain:
    \[
    \Delta_{\mathrm{BIC}}(e)
    = 
    \mathrm{BIC}(X_j\mid\mathcal{P}_j\cup\{X_i\})
    -
    \mathrm{BIC}(X_j\mid\mathcal{P}_j)
    \]

    \State Compute residuals
    $r_j^{-e}$ using $\mathcal{P}_j$
    and $r_j^{+e}$ using
    $\mathcal{P}_j\cup\{X_i\}$.

    \State Residualize $X_i$ with respect to
    $\mathcal{P}_j\setminus\{X_i\}$ to obtain
    $\widetilde{X}_i$.

    \State Compute HSIC gain:
    \[
    \Delta_{\mathrm{HSIC}}(e)
    =
    \mathrm{HSIC}(\widetilde{X}_i,r_j^{-e})
    -
    \mathrm{HSIC}(\widetilde{X}_i,r_j^{+e}).
    \]

    \State Retrieve
    $\mathrm{Prior}_{\mathrm{LLM}}(e)$.

\EndFor
\State Standardize $\Delta\mathrm{BIC}(e)$, $\Delta\mathrm{HSIC}(e)$, and $\mathrm{Prior}_{\mathrm{LLM}}(e)$ across all $e \in E_{\mathrm{disputed}}$.
\State Compute the combined score
\[
S_{\mathrm{combined}}(e)
=
\lambda_{\mathrm{BIC}}(e)
+
\beta_{\mathrm{HSIC}}(e)
+
\gamma_{\mathrm{LLM}}(e)
\]

\State Sort $\mathcal{E}_{\mathrm{disputed}}$ by decreasing $S_{\mathrm{combined}}(e)$.

\ForAll{$e=(X_i\rightarrow X_j)$ in sorted order}

    \If{$X_i=X_j$}
        \State Reject $e$.
        \ElsIf
        {$\Delta_{\mathrm{S}_{\mathrm{combined}}}(e)\leq 0$}
        \State Reject $e$.
    \ElsIf{$\mathcal{E}\cup\{e\}$
        contains a directed cycle}
        \State Reject $e$.
    \Else
        \State $\mathcal{E}
        \gets \mathcal{E}\cup\{e\}$
    \EndIf

\EndFor

\State \Return
$G=(\mathcal{V},\mathcal{E})$

\end{algorithmic}
\end{algorithm}
\section{Case Study of CP Dataset}
\label{sec: case study}
Figure~\ref {fig::Comparison} shows the single-run results of two graphs from the data-driven and LLM-driven causal discovery methods for the Cerebral Palsy dataset. This dataset contains variables that can directly or indirectly affect the metabolic power of children with atypical gait. The data-driven graph misses many true edges (in black), whereas the LLM-driven graph overconnects, for example, connecting demographic variables (age, sex) to all biomechanical variables, such as GDI (in red). In Figure~\ref{fig:final graph}, the graph shows fewer incorrect edges connecting to biomechanical variables and displays the variables with correctly recovered true-positive edges in green. 
\section{LLM Usage}
During manuscript preparation, LLM-based assistance was used for grammar correction, writing refinement, and clarity improvement. They were not used to generate datasets, results, or final scientific conclusions. All research concepts, ideas, and analyses were developed and conducted by the authors.
\section{Prompts}

\begin{figure*}[t]
\centering
\begin{tcolorbox}
    [colback=green!5!white,
    colframe=green!75!black,
    width=\textwidth,
    boxrule=0.6pt,
    arc = 1mm]
\small
\ttfamily
\subsection{No Metadata}

You are an expert causal-discovery assistant.\\
Your task is to judge whether each proposed directed relationship X -> Y is causally plausible.\\
Important rules:\\
1. Use only the variable names and candidate pairs provided.\\
2. Do not assume access to dataset values.\\
3. Judge only the proposed direction X -> Y.\\
4. Do not treat association as causation.\\
5. Be conservative.\\
6. Return STRICT JSON only.\\
7. Do not include markdown.\\
8. Do not include explanations outside JSON.\\
For each pair, decide whether the directed causal relationship is plausible or not.\\

 \label{fig: No metadata}   
\end{tcolorbox}
    
\end{figure*}
\begin{figure*}[t]
\centering
\begin{tcolorbox}
    [colback=green!5!white,
    colframe=green!75!black,
    width=\textwidth,
    boxrule=0.6pt,
    arc = 1mm]
\small
\ttfamily
\subsection{With Metadata}
You are an expert causal-discovery assistant.\\
Your task is to judge whether each proposed directed relationship X -> Y is causally plausible.\\
You must use:\\
1. The provided dataset metadata.\\
2. The provided variable names.\\
3. Your prior scientific/domain knowledge.\\
Important rules:\\
1. Use the metadata as the main dataset-specific context.\\
2. Use prior knowledge only to interpret whether X could plausibly cause Y.\\
3. Do not use dataset values.\\
4. Do not invent variables that are not provided.\\
5. Judge only the proposed direction X -> Y.\\
6. Do not treat association as causation.\\
7. Be conservative.\\
8. Return STRICT JSON only.\\
9. Do not include markdown.\\
10. Do not include explanations.\\
11. Do not include reasoning.\\
12. Do not include confidence scores.\\
For each pair, return only one verdict:\\
- Plausible\\
- Not Plausible\\

    \label{fig: With metadata}
\end{tcolorbox}
    
\end{figure*}
\begin{figure*}[t]
\centering
\begin{tcolorbox}
    [colback=green!5!white,
    colframe=green!75!black,
    width=\textwidth,
    boxrule=0.6pt,
    arc = 1mm]
\small
\ttfamily
\subsection{Schema Generator}
\textbf{Input metadata for the variable pair DMC to Metabolic Power.}
\begin{lstlisting}[
caption={},
label={lst:schema-input}
]
{
  "variable_pair": {
    "cause": "DMC",
    "effect": "Metabolic Power",
    "direction": "DMC -> Metabolic Power"
  },
  "metadata": {
    "DMC": {
      "alias": "dynamic motor control",
      "meaning": "The neuromuscular system's ability to coordinate muscle activation patterns efficiently during walking.",
      "measured_as": "Dynamic Motor Control (DMC) evaluates the complexity of muscle activation during gait by analyzing how muscles are coordinated through synergies. It is calculated by determining how much variance in electromyographic signals is explained by one or more synergies relative to typical development."
    },
    "Metabolic Power": {
      "alias": "net metabolic power",
      "meaning": "The metabolic energy consumed during walking minus the metabolic energy consumed at rest, expressed per unit time.",
      "measured_as": "Breath-to-breath oxygen consumption was measured during six minutes of overground barefoot walking and ten minutes of reclining rest. Steady-state oxygen consumption was converted to power, and net metabolic power was calculated by subtracting resting metabolic power from walking metabolic power."
    }
  }
}
\end{lstlisting}

\textbf{Prompt: }
You are an expert schema/ontology generator for scientific causal discovery datasets. Your task is to generate a SIMPLE causal schema for the dataset variables based on the provided domain and metadata.
You can use prior knowledge to infer the schema. This is a closed system that uses no variables other than the dataset. The main goal of this schema is to help LLMs describe the causal relationships among variables in the next step. That's why the schema should be scientifically sound, categorize variables by their causal nature, and define the role each variable plays in the dataset. How much does its presence matter in the causal diagram?\\
Moreover, the following are critical rules to follow when generating the schema.\\
CRITICAL RULES:\\
1. Return STRICT JSON only.\\
2. Do not include markdown.\\
3. Do not include explanations outside JSON.\\
4. Use categories that are shared by multiple variables whenever possible, and these categories should be causal, not according to the variable's datatype.\\
5. Do NOT create overly specific categories.\\
6. If a dataset has very few variables and one category must contain only one variable, explain why in schema-quality-warnings.\\
7. Categories must support causal reasoning, not just statistical typing.\\
8. Prefer broad, reusable categories.\\
9. Do not create fragmented categories.\\
10. Identifier variables should be marked as identifier and excluded from causal reasoning.\\
11. Outcome variables can be children/effects but should rarely be parents unless scientifically justified.\\

\label{fig:schema generator}
\end{tcolorbox}
    
\end{figure*}
\begin{figure*}[t]
\centering
\begin{tcolorbox}
    [colback=green!5!white,
    colframe=green!75!black,
    width=\textwidth,
    boxrule=0.6pt,
    arc = 1mm]
\small
\ttfamily
12. Be conservative. If uncertain, use unknown and explain briefly in notes.\\
The schema must help later LLM components:\\
- find pairwise causal relationships\\
- verify whether proposed relationships obey schema rules\\
- detect forbidden or implausible causal directions\\
\textbf{Example output from Schema-generator.}
\begin{lstlisting}[
caption={},
label={lst:schema-output}
]
{
  "variables": [
    {
      "name": "DMC",
      "alias": "dynamic motor control",
      "description": "The neuromuscular system's ability to coordinate muscle activation patterns efficiently during walking.",
      "measured_as": "Dynamic Motor Control (DMC) evaluates the complexity of muscle activation during gait by analyzing how muscles are coordinated through synergies.",
      "data_type": "text",
      "semantic_category": "neuromuscular control",
      "causal_role": "intermediate",
      "can_be_parent": true,
      "can_be_child": true,
      "exclude_from_causal_reasoning": false,
      "notes": "Represents a neuromuscular assessment score."
    },
    {
      "name": "Metabolic Power",
      "alias": "net metabolic power",
      "description": "The physiological energy expended during walking after accounting for resting energy expenditure.",
      "measured_as": "Net metabolic power is calculated by subtracting resting metabolic power from walking metabolic power.",
      "data_type": "continuous",
      "semantic_category": "physiological measurement",
      "causal_role": "intermediate",
      "can_be_parent": true,
      "can_be_child": true,
      "exclude_from_causal_reasoning": false,
      "notes": "Measures physiological energy expenditure during gait."
    }
  ],
  "causal_rules": [
    {
      "rule_id": "6",
      "from_category": "neuromuscular control",
      "to_category": "physiological_measurement",
      "relationship": "plausible",
      "reason": "neuromuscular control can influence physiological measurements."
    }
  ]
}
\end{lstlisting}

\end{tcolorbox}
    
\end{figure*}
\begin{figure*}[t]
\centering
\begin{tcolorbox}
    [colback=green!5!white,
    colframe=green!75!black,
    width=\textwidth,
    boxrule=0.6pt,
    arc = 1mm]
\small
\ttfamily
\subsection{Response Generator}
\textbf{Input:\\}
\textbf{Variable pair}:DMC $\rightarrow$ Metabolic Power\\
\& 
\textbf{Metadata}\\
\&
\textbf{Schema} from Schema Generator\\
\textbf{Prompt}:
You are an expert causal reasoning assistant for scientific datasets. Your task is to judge whether a proposed ordered causal relationship X -> Y is plausible. For each pair, generate a response and a one-line reason to choose that response.
Use metadata, schema, and domain knowledge to generate responses.\\

CRITICAL RULES:\\
1. Return STRICT JSON only.\\
2. Do not include markdown.\\
3. Do not include explanations outside JSON.\\
4. Judge only the proposed direction: cause-variable to effect-variable.\\
5. Do not assume association means causation.\\
6. Use the dataset metadata, variable meanings, and generated schema.\\
7. Use schema rules as constraints.\\
8. If the schema states a variable is parent-only, it should not be treated as an effect unless justified by domain knowledge\\
9. If a variable is an identifier, it should not participate in causal reasoning.\\
10. Be conservative.\\
11. If the direction is scientifically weak or unclear, mark it as uncertain or not plausible.\\
12. Always provide a one-line reason as exactly one concise sentence.\\
13. The one-line reason must explain why the edge is plausible, not plausible, uncertain, or reversed.\\
14. Do not invent unsupported domain facts.\\
You must produce JSON matching the required schema.\\
\textbf{Example output from the response generator.}
\begin{lstlisting}[
caption={},
label={lst:response-generator-output}
]
{
  "response_id": 1,
  "cause_variable": "DMC",
  "effect_variable": "Metabolic Power",
  "plausible": true,
  "judgment": "plausible directed cause",
  "confidence": 0.8,
  "one_line_reason": "Dynamic motor control can influence metabolic power by affecting muscle coordination during walking.",
  "schema_consistency": "consistent",
  "schema_rule_used": "rule 6",
  "possible_reverse_edge": false,
  "reverse_edge_reason": "Metabolic power is an outcome measure and does not cause neuromuscular control."
}
\end{lstlisting}
Note: This component produces 9 additional responses through separate API calls using the same prompt configuration.
\label{fig: response generator}
    
\end{tcolorbox}
    
\end{figure*}

\begin{figure*}[t]
\centering
\begin{tcolorbox}
    [colback=green!5!white,
    colframe=green!75!black,
    width=\textwidth,
    boxrule=0.6pt,
    arc = 1mm]
\small
\ttfamily
\subsection{Verifier}
\textbf{Input:\\}
\textbf{Variable pair}:DMC $\rightarrow$ Metabolic Power\\
\textbf{Metadata}\\
\textbf{Schema} from Schema Generator\\
\textbf{Majority Decision }from 10 responses generated by response generator\\
\textbf{Prompt}:
You are an expert causal-relationship verifier. Your task is to verify one directed candidate causal pair X -> Y.\\
You will receive:\\
1. variable-specific schema information\\
2. variable-specific metadata information\\
3. previous majority decision\\
4. previous LLM response counts\\
5. previous rationales if available\\
You must decide whether the previous majority decision should be kept or corrected.\\
Core rules:\\
1. Use the generated schema as the main rulebook.\\
   - The schema may contain variable categories.\\
   - Category-level rules only tell whether relationships between types of variables are generally plausible or implausible.\\
   - Category compatibility supports context only.\\
   - Category compatibility alone is NOT enough to keep or add an edge.\\
2. Use metadata to understand the actual meaning of each variable.\\
   - Metadata may contain variable descriptions and measurement types.\\

3. Do NOT connect all variables from one category to all variables of another category\\
   - Even if schema says category A can affect category B, you must verify the specific variable pair X -> Y.\\
   - You may mark X -> Y plausible only if there is a variable-specific causal reason.\\

4. Keep or add an edge only when:
   - X -> Y does not violate schema\\
   - X -> Y does not violate metadata\\
   - X -> Y has a pair-specific causal mechanism or strong domain-knowledge rationale.\\
5. Remove or reject an edge when:
   - X or Y is excluded from causal reasoning\\
   - X -> Y violates a forbidden edge rule\\
   - X -> Y contradicts metadata\\
   - X -> Y is temporally impossible\\
   - X -> Y is biologically, medically, logically, or mechanistically implausible\\
   - the only support is category-level compatibility\\

6. If schema/metadata do not clearly support or forbid the edge:\\
   - use domain knowledge cautiously.\\
   - If evidence is weak or ambiguous, choose uncertain.\\
   - Do not force plausible.\\

7. Preserve directionality.\\
   - Verify exactly X -> Y.\\
   - Do not reverse the edge.\\
   - If Y to X is plausible but X to Y is not, mark X to Y as not-plausible or uncertain.\\

8. Previous majority decision:\\
   - You may keep it if schema, metadata, and domain knowledge support it.\\
   - You must correct it if it contradicts schema, metadata, or strong domain knowledge.\\
   - You may change uncertain/not-plausible to plausible only if there is a strong pair-specific causal reason.\\

9. Output must be strict JSON only.\\
   - No markdown.\\
   - No explanation outside JSON.\\

    \label{fig: verifier}
\end{tcolorbox}
    
\end{figure*}
\begin{figure*}[t]
\centering
\begin{tcolorbox}
    [colback=green!5!white,
    colframe=green!75!black,
    width=\textwidth,
    boxrule=0.6pt,
    arc = 1mm]
\small
\ttfamily
Decision labels:\\
- plausible\\
- not plausible\\
- uncertain\\
Edge actions:\\
- keep-edge: previous decision was plausible and corrected outcome is plausible.\\
- remove-edge: previous decision was plausible but corrected outcome is not plausible or uncertain.\\
- add-edge: previous decision was not-plausible, uncertain, or no-majority, but corrected outcome is plausible.\\
- no-edge: previous decision was not-plausible, uncertain, or no-majority, and corrected outcome remains not-plausible or uncertain.\\
A valid plausible decision must include a variable-specific reason, such as:\\
- biological mechanism,\\
- clinical pathway,\\
- temporal order,\\
- experimental design,\\
- known functional relationship,\\
- measurement/definition-based dependency.\\
\textbf{Example output from the verifier.}
\begin{lstlisting}[
caption={},
label={}
]
{
  "pair_index": 2,
  "cause_variable": "DMC",
  "effect_variable": "Metabolic Power",
  "direction": "DMC -> Metabolic Power",
  "previous_decision": "plausible",
  "schema_consistency": "consistent",
  "metadata_consistency": "consistent",
  "domain_knowledge_support": "supports",
  "category_level_support_only": false,
  "pair_specific_reason_available": true,
  "contradiction_to_schema_or_metadata": false,
  "contradiction_to_domain_knowledge": false,
  "contradiction_type": "none",
  "corrected_outcome": "plausible",
  "edge_action": "keep-edge",
  "schema_rule_used": "Neuromuscular control can influence physiological measurements.",
  "metadata_evidence_used": "DMC evaluates muscle coordination during walking, which can affect energy expenditure.",
  "domain_knowledge_reason": "Dynamic motor control affects muscle coordination, which can influence energy expenditure and metabolic power during walking.",
  "one_line_reason": "DMC influences muscle coordination and thereby affects walking-related metabolic power.",
  "confidence": 0.95
}
\end{lstlisting}

    \label{fig: output_verifier}
\end{tcolorbox}
    
\end{figure*}

\end{document}